\documentclass[10pt,twocolumn,letterpaper]{article}

\usepackage{iccv}
\usepackage{times}
\usepackage{epsfig}
\usepackage{graphicx}
\usepackage{amsmath}
\usepackage{amssymb}
\usepackage{bm}
\usepackage{dblfloatfix}

\usepackage{multirow}
\usepackage[accsupp]{axessibility}
\usepackage[linesnumbered, ruled,vlined]{algorithm2e}

\usepackage[pagebackref=true,breaklinks=true,letterpaper=true,colorlinks,bookmarks=false]{hyperref}

\iccvfinalcopy 

\def\iccvPaperID{10780} 

\ificcvfinal\fi

\DeclareMathOperator{\tr}{tr}
\DeclareMathOperator*{\argmax}{arg\,max}

\begin{document}

\title{Hierarchical Kinematic Probability Distributions for 3D Human Shape and Pose Estimation from Images in the Wild}

\author{Akash Sengupta\\
University of Cambridge\\
{\tt\small as2562@cam.ac.uk}
\and
Ignas Budvytis\\
University of Cambridge\\
{\tt\small ib255@cam.ac.uk}
\and 
Roberto Cipolla\\
University of Cambridge\\
{\tt\small rc10001@cam.ac.uk}
}

\maketitle
\ificcvfinal\thispagestyle{empty}\fi

\begin{abstract}
This paper addresses the problem of 3D human body shape and pose estimation from an RGB image. This is often an ill-posed problem, since multiple plausible 3D bodies may match the visual evidence present in the input - particularly when the subject is occluded. Thus, it is desirable to estimate a distribution over 3D body shape and pose conditioned on the input image instead of a single 3D reconstruction. We train a deep neural network to estimate a hierarchical matrix-Fisher distribution over relative 3D joint rotation matrices (i.e. body pose), which exploits the human body's kinematic tree structure, as well as a Gaussian distribution over SMPL body shape parameters. To further ensure that the predicted shape and pose distributions match the visual evidence in the input image, we implement a differentiable rejection sampler to impose a reprojection loss between ground-truth 2D joint coordinates and samples from the predicted distributions, projected onto the image plane. We show that our method is competitive with the state-of-the-art in terms of 3D shape and pose metrics on the SSP-3D and 3DPW datasets, while also yielding a structured probability distribution over 3D body shape and pose, with which we can meaningfully quantify prediction uncertainty and sample multiple plausible 3D reconstructions to explain a given input image. Code is available at \url{https://github.com/akashsengupta1997/HierarchicalProbabilistic3DHuman}. 
\end{abstract}
\vspace{-0.2in}

\section{Introduction}
\begin{figure}[t]
 \includegraphics[width=\columnwidth]{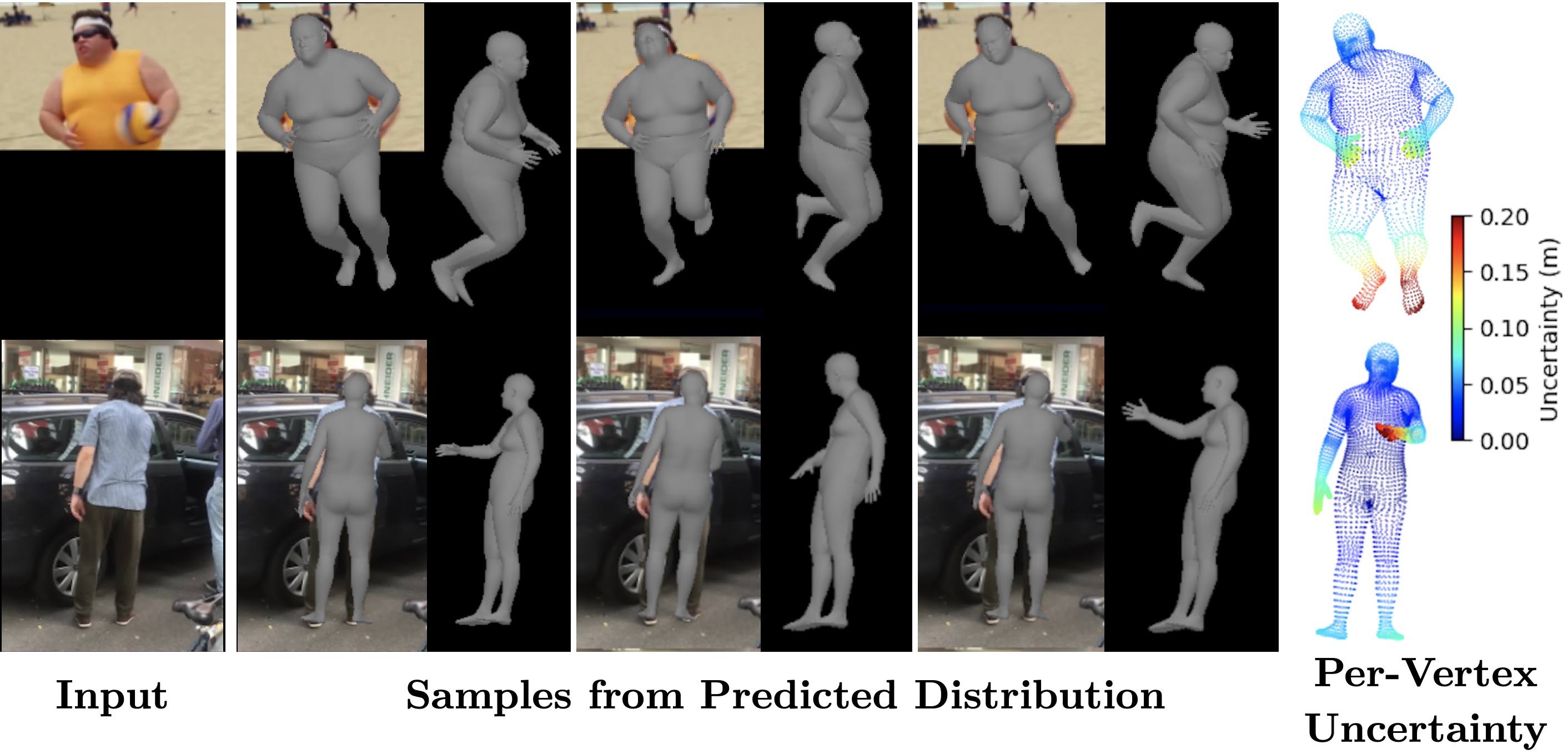}
 \caption{3D reconstruction samples and per-vertex uncertainty corresponding to the predicted hierarchical shape and pose distributions computed from the given input images.}
 \vspace{-0.1in}
 \label{fig:intro}
\end{figure}

3D human body shape and pose estimation from an RGB image is a challenging computer vision problem, partly due to its under-constrained nature wherein multiple 3D human bodies may explain a given 2D image, especially when the subject is significantly occluded, as is common for in-the-wild images. Several recent works \cite{tan2017, hmrKanazawa17, kolotouros2019cmr, kolotouros2019spin, STRAPS2020BMVC, zhang2019danet, georgakis2020hkmr, omran2018nbf, Guler_2019_CVPR_holopose, pavlakos2018humanshape, pavlakos2019texturepose, varol18_bodynet, Moon_2020_ECCV_I2L-MeshNet, smith20193dfromsilhouettes} use deep neural networks to regress a single body shape and pose solution, which can result in impressive 3D body reconstructions given sufficient visual evidence in the input image. However, when visual evidence of the subject's shape and pose is obscured, e.g. due to occluding objects or self-occlusions, a single solution does not fully describe the space of plausible 3D reconstructions. In contrast, we aim to estimate a \textit{structured probability distribution} over 3D body shape and pose, conditioned on the input image, thereby allowing us to sample any number of plausible 3D reconstructions and quantify prediction uncertainty over the 3D body surface, as shown in Figure \ref{fig:intro}.

We use the SMPL body model \cite{SMPL:2015} to represent human shape and pose. Identity-dependent body shape is parameterised by coefficients of a PCA basis - hence, a simple multivariate Gaussian distribution over the shape parameters is suitable. Body pose is parameterised by relative 3D joint rotations along the SMPL kinematic tree, which may be represented using rotation matrices. Regressing rotation matrices using neural networks is non-trivial, since they lie in $SO(3)$, a non-linear 3D manifold with a different topology to $\mathbb{R}^{3 \times 3}$ or $\mathbb{R}^9$, the space in which unconstrained neural network outputs lie. However, one can define probability density functions over the Lie group $SO(3)$, such as the matrix-Fisher distribution \cite{mardia_jupp_2000, down1972orientationstatistics, khatri1977vonmisesfisher}, the parameter of which is an element of $\mathbb{R}^{3 \times 3}$ and may be easily regressed with a neural network \cite{mohlin2020matrixfisher}. We propose a hierarchical probability distribution over relative 3D joint rotations along the SMPL kinematic tree, wherein the probability density function of each joint's relative rotation matrix is a matrix-Fisher distribution conditioned on the parents of that joint in the kinematic tree. We train a deep neural network to predict the parameters of such a distribution over body pose, alongside a Gaussian distribution over SMPL shape.

Moreover, to ensure that 3D bodies sampled from the predicted distributions match the 2D input image, we implement a reprojection loss between predicted samples and ground-truth visible 2D joint annotations. To allow for the backpropagation of gradients through the sampling operation, we present a differentiable rejection sampler for matrix-Fisher distributions over relative 3D joint rotations.

Finally, a key obstacle for SMPL body shape regression from in-the-wild images is the lack of training datasets with \textit{accurate} and \textit{diverse} body shape labels \cite{STRAPS2020BMVC}. To overcome this, we follow \cite{STRAPS2020BMVC, smith20193dfromsilhouettes, pavlakos2018humanshape, sengupta2021probabilisticposeshape} and utilise synthetic data, randomly generated on-the-fly during training. Inspired by \cite{Charles2020realtimesscreen}, we use convolutional edge filters to close the large synthetic-to-real gap and show that using edge-based inputs yields better performance than commonly-used silhouette-based inputs \cite{STRAPS2020BMVC, smith20193dfromsilhouettes, sengupta2021probabilisticposeshape, pavlakos2018humanshape}, due to improved robustness and capacity to retain visual shape information.

In summary, our main contributions are as follows:
\begin{itemize}
    \item Given an input image, we predict a novel hierarchical matrix-Fisher distribution over relative 3D joint rotation matrices, whose structure is explicitly informed by the SMPL kinematic tree, alongside a Gaussian distribution over SMPL shape parameters. 
    
    \item We present a differentiable rejection sampler to sample any number of plausible 3D reconstructions and quantify prediction uncertainty over the body surface. This enables a reprojection loss between predicted samples and ground-truth coordinates of visible 2D joints, further ensuring that the predicted distributions are consistent with the input image.
    
    \item We use simple convolutional edge filters to improve the random synthetic training framework used by \cite{STRAPS2020BMVC, sengupta2021probabilisticposeshape}. Edge filtering is a computationally-cheap and robust method for closing the domain gap between synthetic RGB training data and real RGB test data.
\end{itemize}
 \vspace{-0.1in}

\begin{figure*}[t]
 \centering
 \includegraphics[width=\textwidth]{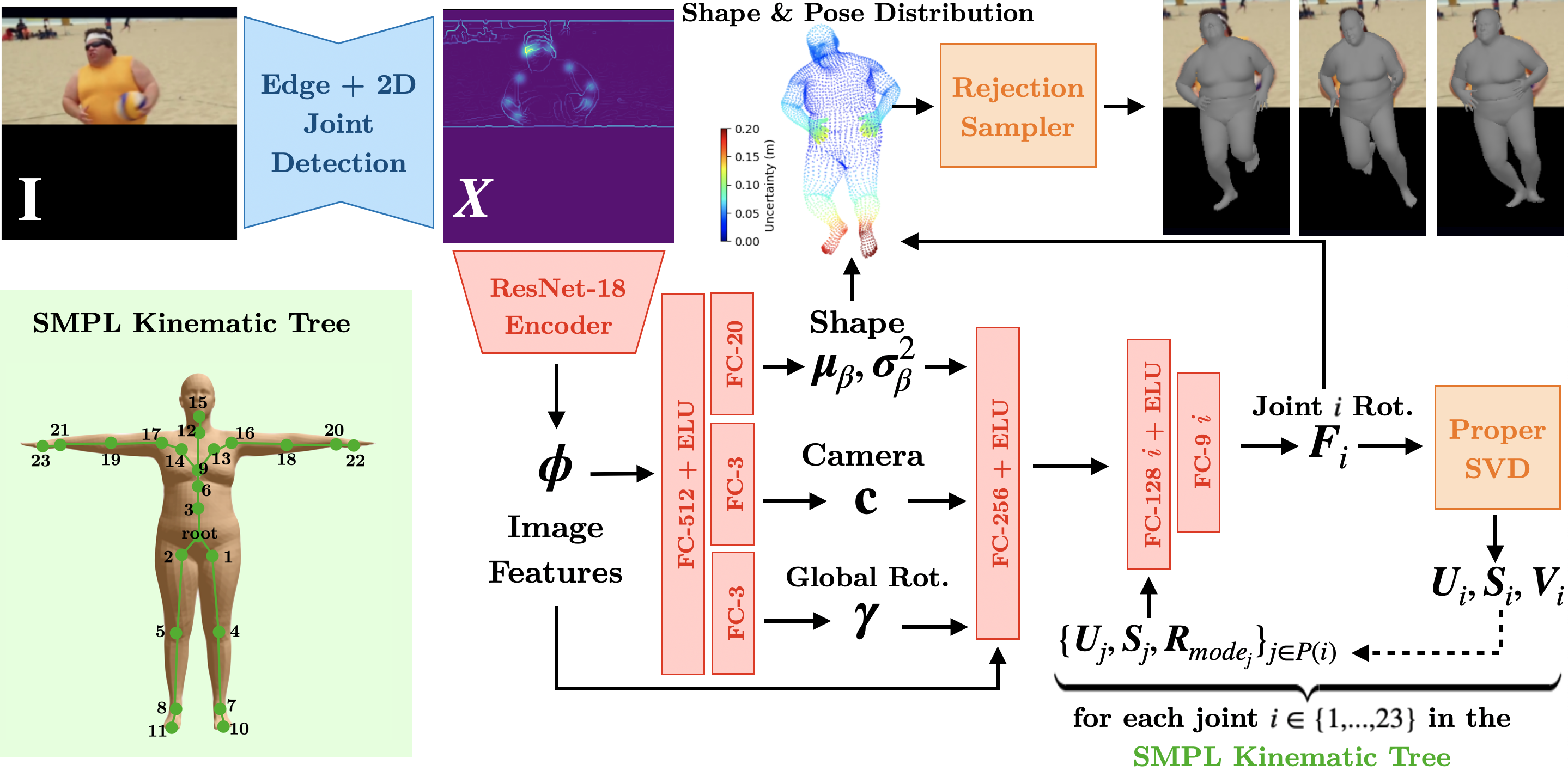}
 \caption{Network architecture of our hierarchical SMPL \cite{SMPL:2015} shape and pose distribution predictor. The input image is converted into an edge-and-joint-heatmap proxy representation, which is passed through the prediction network to produce distributions over shape parameters and relative 3D joint rotation matrices. Rejection sampling is used to sample 3D reconstructions from the predicted distributions.}
 \label{fig:method}
\end{figure*}

\section{Related Work}

This section reviews approaches to monocular 3D human body shape and pose estimation, as well as deep-learning-based methods for probabilistic rotation estimation.

\noindent \textbf{Monocular 3D shape and pose estimation} methods can be classified as optimisation-based or learning-based. Optimisation-based approaches fit a parametric 3D body model \cite{SMPL:2015, Anguelov05scape:shape, SMPL-X:2019, Joo_2018_CVPR_total_capture} to 2D observations, such as 2D keypoints \cite{Bogo:ECCV:2016, Lassner:UP:2017}, silhouettes \cite{Lassner:UP:2017} or body part segmentations \cite{Zanfir_2018_CVPR}, by optimising a suitable cost function. These methods do not require expensive 3D-labelled training data, but are sensitive to poor intialisations and noisy observations. 

Learning-based approaches can be further split into model-free or model-based. Model-free methods use deep networks to directly output human body vertex meshes \cite{kolotouros2019cmr, Moon_2020_ECCV_I2L-MeshNet, zhang2019danet, Zeng_2020_CVPR_mesh_dense, Choi_2020_ECCV_Pose2Mesh}, voxel grids \cite{varol18_bodynet} or implicit surfaces \cite{saito2019pifu, saito2020pifuhd} from an input image. In contrast, model-based methods \cite{hmrKanazawa17, STRAPS2020BMVC, omran2018nbf, georgakis2020hkmr, tan2017, Guler_2019_CVPR_holopose, pavlakos2018humanshape, pavlakos2019texturepose, Xu_2019_ICCV} regress 3D body model parameters \cite{SMPL-X:2019, SMPL:2015, Joo_2018_CVPR_total_capture, Anguelov05scape:shape}, which give a low-dimensional representation of a 3D human body. To overcome the lack of in-the-wild 3D-labelled training data, several methods \cite{hmrKanazawa17, Xu_2019_ICCV, kolotouros2019cmr, georgakis2020hkmr, Guler_2019_CVPR_holopose} use diverse 2D-labelled data as a source of weak supervision. \cite{kolotouros2019spin} extends this approach by incorporating optimisation into their model training loop, lifting 2D labels to self-improving 3D labels. These approaches often result in impressive 3D pose predictions, but struggle to accurately predict a diverse range of body shapes, since 2D keypoint supervision only provides a sparse shape signal. Shape prediction accuracy may be improved using synthetic training data \cite{STRAPS2020BMVC, smith20193dfromsilhouettes, pavlakos2018humanshape, sengupta2021probabilisticposeshape} consisting of synthetic input proxy representations (PRs) paired with ground-truth body shape and pose. PRs commonly consist of silhouettes and 2D joint heatmaps \cite{STRAPS2020BMVC, pavlakos2018humanshape, sengupta2021probabilisticposeshape}, necessitating accurate silhouette segmentations \cite{kirillov2019pointrend, Guler2018DensePose} at test-time, which is not guaranteed for challenging in-the-wild inputs. Other methods \cite{varol18_bodynet} pre-train on synthetic RGB inputs \cite{varol17_surreal} and then fine-tune on the scarce and limited-shape-diversity real 3D training data available \cite{h36m_pami, vonMarcard2018}, to avoid over-fitting to artefacts in low-fidelity synthetic data. In contrast, we utilise edge-based PRs, hence dropping the reliance on accurate segmentation networks without requiring fine-tuning on real data or high-fidelity synthetic data.

\noindent \textbf{3D human shape and pose distribution estimation.} Early optimisation-based 3D pose estimators \cite{Sminchisescu2001covsampling, Sminchisescu2002hyper, Sminchisescu2003kinematicjump, Choo2001tracking, Deutscher2000particle} specified a cost function corresponding to the posterior probability of 3D pose given 2D observations and analysed its multi-modal structure due to ill-posedness. Strategies to sample multiple 3D poses with high posterior probability included cost-covariance-scaled \cite{Sminchisescu2001covsampling} and inverse-kinematics-based \cite{Sminchisescu2003kinematicjump} global search and local refinement, as well as cost-function-modifying MCMC \cite{Sminchisescu2002hyper}. Recently, several learning-based methods \cite{Sminchisescu2005bm3e, Li_2019_CVPR, Jahangiri2017ICCVW, Wehrbein2021posenormflows, Oikarinen2020graphmdn} predict multi-modal distributions over 3D joint locations conditioned on 2D inputs, using Bayesian mixture of experts \cite{Sminchisescu2005bm3e}, mixture density networks \cite{Li_2019_CVPR,Bishop94mixturedensity, Oikarinen2020graphmdn} or normalising flows \cite{Wehrbein2021posenormflows, Rezende2015normflows}. Our method extends beyond 3D joints and predicts distributions over human pose \textit{and shape}. This has been addressed by Biggs \etal \cite{biggs2020multibodies}, who predict a categorical distribution over a set of SMPL \cite{SMPL:2015} parameter hypotheses. Sengupta \etal \cite{sengupta2021probabilisticposeshape} estimate an independent Gaussian distribution over both SMPL shape and joint rotation vectors. In contrast, we note that 3D rotations lie in $SO(3)$, motivating our hierarchical matrix-Fisher distribution. 

\noindent \textbf{Rotation distribution estimation via deep learning.} Prokudin \etal \cite{deepdirectstat2018} use biternion networks to predict a mixture-of-von-Mises distribution over object pose angle. Gilitschenski \etal \cite{Gilitschenski2020} use a Bingham distribution over unit quaternions to represent orientation uncertainty. However, these works have to enforce constraints on the parameters of their predicted distributions (e.g. positive semi-definiteness). To overcome this, Mohlin \etal \cite{mohlin2020matrixfisher} train a deep network to regress a matrix-Fisher distribution \cite{mardia_jupp_2000, down1972orientationstatistics, khatri1977vonmisesfisher} over 3D rotation matrices. We adapt this approach to define our hierarchical matrix-Fisher distribution over relative 3D joint rotation matrices.

\section{Method}
\label{sec:method}

This section provides an overview of SMPL \cite{SMPL:2015} and the matrix-Fisher distribution \cite{down1972orientationstatistics, khatri1977vonmisesfisher, mardia_jupp_2000}, presents our structured, hierarchical pose and shape distribution estimation architecture and discusses the loss functions used to train it.

\subsection{SMPL model}
SMPL \cite{SMPL:2015} is a parametric 3D human body model. Identity-dependent body shape is represented by shape parameters $\boldsymbol{\beta} \in \mathbb{R}^{10}$, which are coefficients of a PCA body shape basis. Body pose is defined by the relative 3D rotations of the bones formed by the 23 body (i.e. non-root) joints in the SMPL kinematic tree. The rotations may be represented using rotation matrices $\{\bm{R}_i\}_{i=1}^{23}$, where $\bm{R}_i \in SO(3)$. We parameterise the global rotation (i.e. rotation of the root joint) in axis-angle form by $\boldsymbol{\gamma} \in \mathbb{R}^3$. A differentiable function $\mathcal{S}(\{\bm{R}_i\}_{i=1}^{23}, \boldsymbol{\beta}, \boldsymbol{\gamma})$ maps the input pose and shape parameters to an output vertex mesh $\bm{V} \in \mathbb{R}^{6890 \times 3}$. 3D joint locations, for $L$ joints of interest, are obtained as $\bm{J}^{\text{3D}} = \mathcal{J}\bm{V}$ where $\mathcal{J} \in \mathbb{R}^{L \times 6890}$ is a linear vertex-to-joint regression matrix.

\subsection{Matrix-Fisher distribution over $SO(3)$}
The 3D special orthogonal group may be defined as $SO(3) = \{\bm{R} \in \mathbb{R}^{3\times3} | \bm{R}^T\bm{R} = \bm{I}, \text{det}(\bm{R}) = 1\}$. The matrix-Fisher distribution \cite{down1972orientationstatistics, khatri1977vonmisesfisher, mardia_jupp_2000} defines a probability density function over $SO(3)$, given by
\begin{equation}
    p(\bm{R}|\bm{F}) = \frac{1}{c(\bm{F})}\exp(\tr(\bm{F}^T\bm{R})) = \mathcal{M}(\bm{R}; \bm{F})
\label{eqn:mf_pdf}
\end{equation}
where $\bm{F} \in \mathbb{R}^{3 \times 3}$ is the matrix parameter of the distribution, $c(\bm{F})$ is the normalising constant and $\bm{R} \in SO(3)$. We present some key properties of the matrix-Fisher distribution below, but refer the reader to \cite{lee2018bayesianattitude, mohlin2020matrixfisher} for further details, visualisations and a method for approximating the intractable normalising constant and its gradient w.r.t. $\bm{F}$.

The properties of $\mathcal{M}(\bm{R}; \bm{F})$ can be described in terms of the singular value decomposition (SVD) of $\bm{F}$, denoted by $\bm{F} = \bm{U}'\bm{S}'\bm{V}'^T$, with $\bm{S}' = \text{diag}(s'_1, s'_2, s'_3)$. $\bm{U}'$ and $\bm{V}'$ are orthonormal matrices, but they may have a determinant of -1 and thus are not necessarily elements of $SO(3)$. Therefore, a \textit{proper} SVD \cite{lee2018bayesianattitude} $\bm{F}=\bm{U}\bm{S}\bm{V}^T$ is used, where
\begin{equation}
\begin{aligned}
\bm{U} &= \bm{U}' \text{diag}(1, 1, \det(\bm{U}'))\\
\bm{V} &= \bm{V}' \text{diag}(1, 1, \det(\bm{V}'))\\
\bm{S} &= \text{diag}(s_1, s_2, s_3) = \text{diag}(s'_1, s'_2, \det(\bm{U}'\bm{V}')s'_3)\\
\end{aligned}
\label{eqn:proper_svd}
\vspace{-0.1cm}
\end{equation}
which ensures that $\bm{U}, \bm{V} \in SO(3)$. Then, the \textit{mode} of the distribution is given by \cite{lee2018bayesianattitude}
\begin{equation}
    \bm{R}_{\text{mode}} = \argmax_{\bm{R} \in SO(3)} p(\bm{R}|\bm{F}) = \bm{U}\bm{V}^T.  
\label{eqn:mf_mode}
\end{equation}
The columns of $\bm{U}$ define the distribution's \textit{principal axes} of rotation (analogous to the principal axes of a multivariate Gaussian distribution), while the proper singular values in $\bm{S}$ give the \textit{concentration} of the distribution for rotations about the principal axes \cite{lee2018bayesianattitude}. Specifically, the concentration along rotations of $\bm{R}_\text{mode}$ about the $i$-th principal axis ($i$-th column of $\bm{U}$) is given by $s_j + s_k$ for $(i,j,k) \in \{(1, 2, 3), (2, 3, 1), (3, 1, 2)\}$. The concentration of the distribution may be different about each principal axis, allowing for axis-dependent rotation uncertainty modelling.

\subsection{Proxy representation computation}
 Given an input RGB image $\mathbf{I}$, we first compute a proxy representation $\bm{X}$ (see Figure \ref{fig:method}), consisting of an edge-image concatenated with joint heatmaps. Comparisons with silhouette- and RGB-based representations are given in Section \ref{subsec:ablation}. Edge-images are obtained with Canny edge detection \cite{canny1986edge}. 2D joint heatmaps are computed using HRNet-W48 \cite{sun2019hrnet}, and joint predictions with low confidence scores ($<0.6$) are thresholded out. The edge-image and joint heatmaps are stacked along the channel dimension to produce $\bm{X} \in \mathbb{R}^{H \times W \times (L+1)}$. Proxy representations \cite{STRAPS2020BMVC,pavlakos2018humanshape} are used to close the domain gap between synthetic training images and real test-time RGB images, since synthetic proxy representations are more similar to their real counterparts than synthetic RGB images are to real RGB images.

\subsection{Body shape and pose distribution prediction}
\label{subsec:distribution_predictor}
Our goal is to predict a probability distribution over relative 3D joint rotations $\{\bm{R}_i\}_{i=1}^{23}$ and SMPL shape parameters $\boldsymbol{\beta}$ conditioned upon a given input proxy representation $\bm{X}$. We also predict deterministic estimates of the global body rotation $\boldsymbol{\gamma}$ and weak-perspective camera parameters $\mathbf{c} = [s, t_x, t_y]$, representing scale and $xy$ translation.

Since $\boldsymbol{\beta}$ represents the linear coefficients of a PCA shape-space, a Gaussian distribution with a diagonal covariance matrix is suitable \cite{sengupta2021probabilisticposeshape},
\begin{equation}
    p(\bm{\beta} | \bm{X}) = \mathcal{N}(\bm{\beta}; \bm{\mu}_\beta(\bm{X}), \text{diag}(\bm{\sigma}^2_\beta(\bm{X}))
\label{eqn:gaussian_shape}
\end{equation}
where the mean $\bm{\mu}_\beta$ and variances $\bm{\sigma}^2_\beta$ are functions of $\bm{X}$.

The matrix-Fisher distribution (Equation \ref{eqn:mf_pdf}) may be naively used to define a distribution over 3D joint rotations
\begin{equation}
    p(\bm{R}_i | \bm{X}) = \mathcal{M}(\bm{R}_i; \bm{F}_i(\bm{X}))
\label{eqn:indep_mf_joints}
\end{equation}
for $i \in \{1, 2, ..., 23\}$. Here, each joint is modelled \textit{independently} of all the other joints. Thus, the matrix parameter of the $i$-th joint, $\bm{F}_i$, is a function of the input $\bm{X}$ only. 

To predict the parameters of this naive, independent distribution over 3D joint rotations, in addition to the shape distribution parameters, global body rotation and weak-perspective camera, we learn a function $f_\text{indep}$ mapping the input $\bm{X}$ to the set of desired outputs $Y = \{\{\bm{F}_i\}_{i=1}^{23}, \bm{\mu}_\beta, \bm{\sigma}_\beta^2, \bm{\gamma}, \mathbf{c}\}$, where $f_\text{indep}$ is represented by a deep neural network with weights $\mathbf{W}_\text{indep}$.

However, the independent matrix-Fisher distribution in Equation \ref{eqn:indep_mf_joints} does not model SMPL 3D joint rotations faithfully, since the rotation of each part/bone is defined \textit{relative} to its parent joint in the SMPL kinematic tree. Hence, a distribution over the $i$-th rotation matrix $\bm{R}_i$ conditioned on the input $\bm{X}$ should be informed by the distributions over all its parent joints $P(i)$, as well as the global body rotation $\bm{\gamma}$, to enable the distribution to match the 2D visual pose evidence present in $\bm{X}$. Furthermore, 3D joints in the SMPL rest-pose skeleton are dependent upon the shape parameters $\bm{\beta}$, while the mapping from 3D to the 2D image plane is given by the camera model. Hence, a distribution over $\bm{R}_i$ given $\bm{X}$ should also consider the predicted shape mean $\bm{\mu}_\beta$ and variance $\bm{\sigma}_\beta^2$, as well as the predicted camera $\mathbf{c}$. This is similar to the rationale behind the deterministic iterative/hierarchical predictors in \cite{hmrKanazawa17, georgakis2020hkmr}, except we model these relationships in a probabilistic sense, by defining
\begin{equation}
\begin{aligned}
    p\big(\bm{R}_i | \bm{X}, \{\bm{F}_j\}_{j \in P(i)}, \bm{\gamma}, \bm{\mu}_\beta, \bm{\sigma}_\beta^2, \mathbf{c}\big) = \mathcal{M}(\bm{R}_i; \bm{F}_i)\\
    \bm{F}_i = f_i\big(\bm{X}, \{(\bm{U}_j, \bm{S}_j, \bm{R}_{\text{mode}_j})\}_{j \in P(i)}, \bm{\gamma}, \bm{\mu}_\beta, \bm{\sigma}_\beta^2, \mathbf{c}\big)
\end{aligned}
\label{eqn:hier_mf_joints}
\end{equation}
for $i \in \{1, 2, ..., 23\}$. Now, the matrix parameter of the $i$-th joint is a function of all its parent distributions, represented by the principal axes $\bm{U}_j$, singular values $\bm{S}_j$ and modes $\bm{R}_{\text{mode}_j} = \bm{U}_j\bm{V}_j^T$ for $j \in P(i)$, as well as the shape distribution $\{\bm{\mu}_\beta, \bm{\sigma}_\beta^2\}$, global rotation $\bm{\gamma}$, camera parameters $\mathbf{c}$ and the input $\bm{X}$. Note that the parent distributions are themselves functions of \textit{their} respective parent joints, while $\bm{\gamma}, \bm{\mu}_\beta, \bm{\sigma}_\beta^2$ and $\mathbf{c}$ are all functions of $\bm{X}$.

To predict the parameters of the hierarchical matrix-Fisher distribution in Equation \ref{eqn:hier_mf_joints}, we propose a hierarchical neural network architecture $f_\text{hier}$, with weights $\mathbf{W}_\text{hier}$ (Figure \ref{fig:method}). When considered as a black-box, $f_\text{hier}$ yields the same set of outputs $Y$ as $f_\text{indep}$. However, $f_\text{hier}$ utilises the iterative hierarchical architecture presented in Figure \ref{fig:method}, which amounts to multiple streams of fully-connected layers, each following one ``limb'' of the kinematic tree. In contrast, $f_\text{indep}$ predicts pose similarly to shape, camera and global rotation parameters, using a single stream of fully-connected layers. We compare the naive independent formulation with the hierarchical formulation in Section \ref{subsec:ablation}.

\subsection{Loss functions}
\label{subsec:losses}
Distribution prediction networks are trained with a synthetic dataset $\{\bm{X}^n, (\{\bm{R}_i^n\}_{i=1}^{23}, \bm{\beta}^n, \bm{\gamma}^n)\}_{n=1}^N$ (Section \ref{sec:implementation}).

\noindent \textbf{Negative log-likelihood (NLL) loss on distribution parameters.} The NLL corresponding to the Gaussian body shape distribution (Equation \ref{eqn:gaussian_shape}) is given by: 
\begin{equation}
 \mathcal{L}_{\beta\text{-NLL}} = - \sum_{n=1}^N \log \mathcal{N}\big(\bm{\beta}^n; \bm{\mu}_\beta(\bm{X}^n), \text{diag}(\bm{\sigma}^2_\beta(\bm{X}^n))\big).
\label{eqn:shape_nll}
\end{equation}

The NLL corresponding to the matrix-Fisher distribution over relative 3D joint rotations is defined as \cite{mohlin2020matrixfisher}:
\begin{equation}
\begin{aligned}
\mathcal{L}_{R\text{-NLL}} &= - \sum_{n=1}^N \log \mathcal{M}(\bm{R}^n_i; \bm{F}^n_i)\\
&= \sum_{n=1}^N \log c(\bm{F}^n_i) - \tr({\bm{F}_i^{nT} \bm{R}^n_i})
\end{aligned}
\vspace{-0.05in}
\label{eqn:pose_nll}
\end{equation}
for $i \in \{1, 2, ..., 23\}$, where $\bm{F}^n_i$ may be obtained via the independent or hierarchical matrix-Fisher models presented above. Intuitively, the trace term pushes the predicted distribution mode $\bm{R}^n_{\text{mode}_i}$ (Equation \ref{eqn:mf_mode}) towards the target $\bm{R}^n_i$, while the log normalising constant acts as a regulariser, preventing the singular values of $\bm{F}^n_i$ from getting too large \cite{mohlin2020matrixfisher}. All predicted distribution parameters are dependent on the model weights, $\mathbf{W}_\text{indep}$ or $\mathbf{W}_\text{hier}$, which are learnt in a maximum likelihood framework aiming to minimise the joint shape and pose NLL: $\mathcal{L}_{\text{NLL}} = \mathcal{L}_{\beta\text{-NLL}} + \mathcal{L}_{R\text{-NLL}}$.

\noindent \textbf{Loss on global body rotation.} We predict deterministic estimates of the global body rotation vectors $\hat{\bm{\gamma}}^n$, which are supervised using ground-truth global rotations ${\bm{\gamma}}^n$, with loss $\mathcal{L}_{\text{global}} = \sum_{n=1}^N \| \bm{R}(\bm{\gamma}_n) - \bm{R}( \bm{\hat{\gamma}}_n) \|_{F}^2$. $ \bm{R}( \bm{\gamma}) \in SO(3)$ is the rotation matrix corresponding to $\bm{\gamma}$. 

\noindent \textbf{2D joints loss on samples.} Applying $\mathcal{L}_{\text{NLL}}$ alone results in overly uncertain predicted 3D shape and pose distributions (see Section \ref{subsec:ablation}). To ensure that the predicted distributions match the visual evidence in the input $\bm{X}^n$, we impose a reprojection loss between ground-truth 2D joint coordinates (in the image plane) and predicted 2D joint samples, which are obtained by \textit{differentiably sampling} 3D bodies from the predicted distributions and projecting to 2D using the predicted camera $\mathbf{c}^n = [s^n, t_x^n, t_y^n]$. Ground-truth 2D joints $\bm{J}^n_\text{2D}$ are computed from $\{\{\bm{R}^n_i\}_{i=1}^{23}, \bm{\beta}^n, \bm{\gamma}^n\}$ during synthetic training data generation (see Section \ref{sec:implementation}). 

\begin{algorithm}[t]
\SetAlgoLined
\KwIn{$\bm{U}$, $\bm{S}=\text{diag}(s_1, s_2, s_3)$, $\bm{V}$, $b$}
\KwOut{$\hat{\bm{R}} \in SO(3)$ s.t. $\hat{\bm{R}} \sim \mathcal{M}(\bm{R};\bm{U}\bm{S}\bm{V}^T)$}
$\bm{A} = \text{diag}(0, 2(s_2+s_3), 2(s_1+s_3), 2(s_1+s_2))$\\
$\bm{\Omega} = \bm{I}_4 + \frac{2}{b}\bm{A}$\\
$M = \exp\big(\frac{b-4}{2}\big)\big(\frac{4}{b}\big)^2$\\
\Repeat{$w < \frac{\exp(-\bm{x}^T\bm{A}\bm{x})}{M(\bm{x}^T\bm{\Omega}\bm{x})^{-2}}$}{
  Sample $\bm{\epsilon} \sim \mathcal{N}(\mathbf{0}_4, \bm{I}_4$)\\
  $\bm{y} = (\bm{\Omega}^{-1})^\frac{1}{2} \bm{\epsilon}$\\
  Propose $\bm{x} = \frac{\bm{y}}{\|\bm{y}\|}$ s.t. $\bm{x} \in S^3$\\
  Sample $w \sim \text{Unif}[0, 1]$
 }
 $\hat{\bm{Q}} = quaternion\_to\_matrix(\bm{x})$ s.t. $\hat{\bm{Q}} \in SO(3)$\\
 \KwRet{$\hat{\bm{R}} = \bm{U}\hat{\bm{Q}}\bm{V}^T$}\\
 \caption{Differentiable Rejection Sampler}
 \label{alg:rejection}
\end{algorithm}

We adapt the rejection sampler presented in \cite{kent2013binghamsampling} to sample from a matrix-Fisher distribution $\mathcal{M}(\bm{R}; \bm{F})$, modifying it to allow for backpropagation of gradients through the proposal sampling step (lines 5-7 in Algorithm \ref{alg:rejection}). We refer the reader to \cite{kent2013binghamsampling} for further details about the rejection sampler. In short, to simulate a matrix-Fisher distribution with parameter $\bm{F}=\bm{U}\bm{S}\bm{V}^T,$ we sample unit quaternions from a Bingham distribution \cite{mardia_jupp_2000} over the unit 3-sphere $S^3$, with Bingham parameter $\bm{A}$ computed from $\bm{S}$, and then convert the sampled quaternions into rotation matrices \cite{kent2013binghamsampling, mardia_jupp_2000} with the desired matrix-Fisher distribution. Rejection sampling is used to sample from the Bingham distribution, which has pdf $p_\text{Bing}(\bm{x}) \propto \exp(-\bm{x}^T\bm{A}\bm{x})$ for $\bm{x} \in S^3$. The proposal distribution for the rejection sampler is an angular central Gaussian (ACG) distribution, with pdf $p_\text{ACG}(\bm{x}) \propto (\bm{x}^T\bm{\Omega}\bm{x})^{-2}$. The ACG distribution is easily simulated \cite{kent2013binghamsampling} by sampling from a zero-mean Gaussian distribution with covariance matrix $\bm{\Omega}^{-1}$ and normalising to unit-length (lines 5-7 in Algorithm \ref{alg:rejection}). The re-parameterisation trick \cite{kingma2014autoencoding} is used to differentiably sample from this zero-mean Gaussian, thus allowing for backpropagation of gradients through the rejection sampler.

Algorithm \ref{alg:rejection} samples $K$ sets of relative 3D joint rotation matrices $\{\{\hat{\bm{R}}^n_{i,k}\}_{i=1}^{23}\}_{k=1}^{K}$ from the corresponding distributions $\{\mathcal{M}(\bm{R}^n_i; \bm{F}^n_i)\}_{i=1}^{23}$. Furthermore, we differentiably sample $K$ SMPL shape vectors from the predicted Gaussian distribution $\{\hat{\bm{\beta}}^n_k \sim \mathcal{N}(\bm{\beta}; \bm{\mu}_\beta(\bm{X}^n), \text{diag}(\bm{\sigma}^2_\beta(\bm{X}^n)))\}_{k=1}^K$, again using the re-parameterisation trick \cite{kingma2014autoencoding}.

The body shape and 3D joint rotation samples are converted into 2D joint samples using the SMPL model and weak-perspective camera parameters
\begin{equation}
\hat{\bm{J}}^n_{\text{2D}_k} = s^n \Pi(\mathcal{J} \mathcal{S}(\{\hat{\bm{R}}^n_{i,k}\}_{i=1}^{23}, \hat{\bm{\beta}}^n_k, \hat{\bm{\gamma}}^n)) + [t^n_x, t^n_y]
\end{equation}
where $\Pi()$ is an orthographic projection. The reprojection loss applied between the predicted 2D joint samples and the visible target 2D joint coordinates is given by
\begin{equation}
\mathcal{L}_\text{2D Samples} = \sum_{n=1}^{N} \sum_{k=1}^{K} \|\bm{\omega}^n (\bm{J}^n_\text{2D} - \hat{\bm{J}}^n_{\text{2D}_k})\|_2^2 
\label{eqn:2d_loss}
\end{equation}
where the visibilities of the target joints are denoted by $\boldsymbol{\omega}^n \in \{0,1\}^L$  (1 if visible, 0 otherwise).

\begin{figure*}[t]
    \centering
    \includegraphics[width=\textwidth]{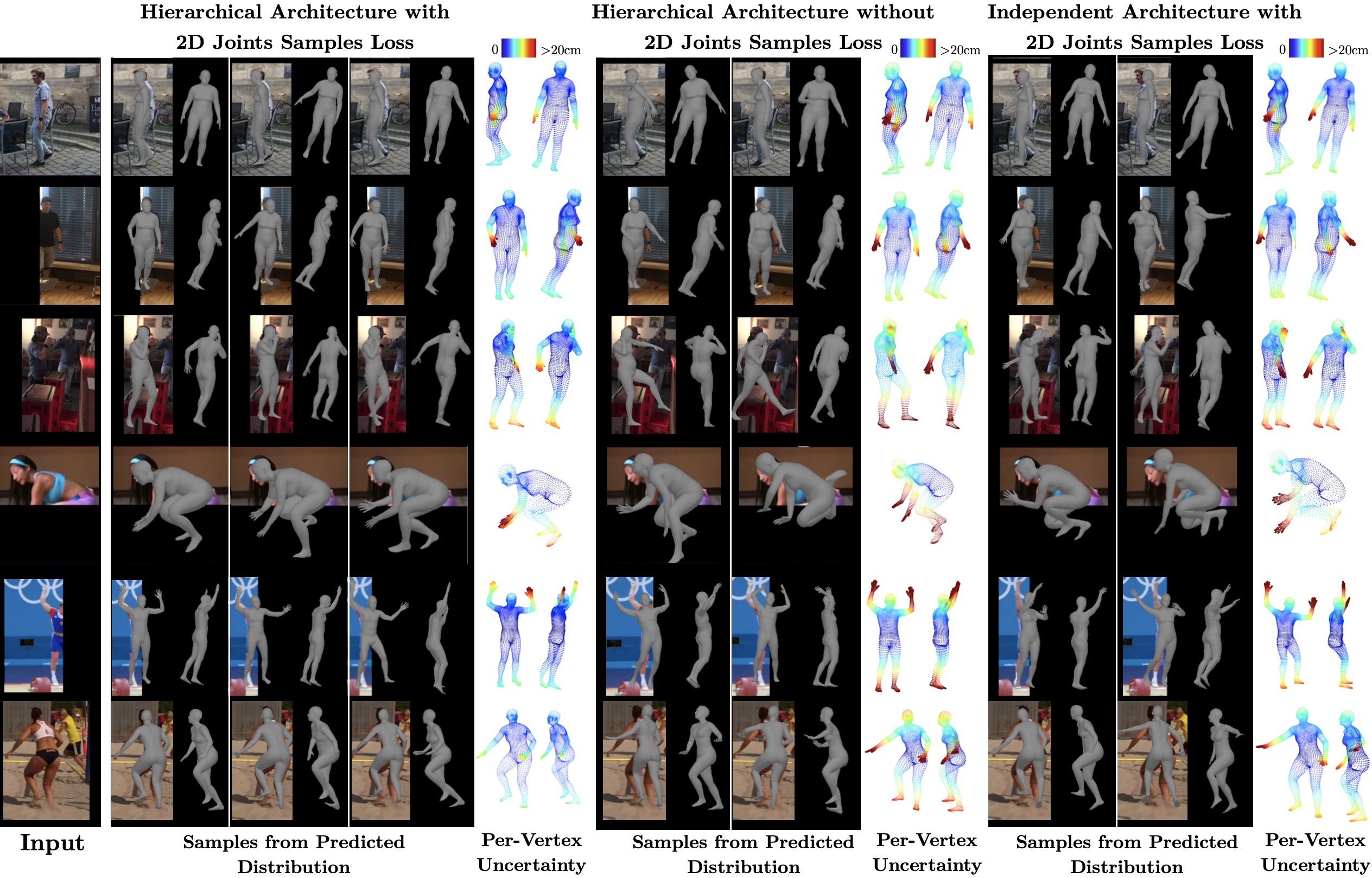}
    \caption{3D reconstruction samples and per-vertex uncertainty corresponding to shape and pose distributions predicted using the hierarchical architecture with 2D samples loss (left), hierarchical architecture without 2D samples loss (centre) and independent architecture with 2D samples loss (right). Per-vertex uncertainty (in cm) is estimated by sampling 100 SMPL meshes from the predicted distributions and determining the average Euclidean distance from the sample mean for each vertex. Both the hierarchical architecture \textit{and} the sample reprojection loss are required for predicted distributions to match the inputs, while demonstrating greater uncertainty for ambiguous parts.}
    \label{fig:ablation}
    \vspace{-0.3cm}
\end{figure*}

\section{Implementation Details}
\label{sec:implementation}

\noindent \textbf{Synthetic training data.} To train our 3D body shape and pose distribution prediction networks, we require a training dataset $\{\bm{X}^n, (\{\bm{R}_i^n\}_{i=1}^{23}, \bm{\beta}^n, \bm{\gamma}^n)\}_{n=1}^N$. We extend the synthetic training frameworks presented in \cite{STRAPS2020BMVC, sengupta2021probabilisticposeshape}, which involve generating inputs and corresponding SMPL body shape and pose (i.e. 3D joint rotation) labels randomly and on-the-fly during training. In brief, for every training iteration, SMPL shapes $\bm{\beta}^n$ are randomly sampled from a prior Gaussian distribution while relative 3D joint rotations $\{\bm{R}_i^n\}_{i=1}^{23}$ and global rotation $\bm{\gamma}^n$ are chosen from the training sets of UP-3D \cite{Lassner:UP:2017}, 3DPW \cite{vonMarcard2018} or Human3.6M \cite{h36m_pami}. These are converted into training inputs $\bm{X}^n$ and ground-truth 2D joint coordinates $\bm{J}^n$ using the SMPL model and a light-weight renderer \cite{ravi2020pytorch3d}. Cropping, occlusion and noise augmentations are then applied to the synthetic inputs.

Previous synthetic training frameworks \cite{STRAPS2020BMVC, sengupta2021probabilisticposeshape, smith20193dfromsilhouettes} often use silhouette-based training inputs. This necessitates accurate human silhouette segmentation at test-time, which may be challenging to do robustly. In contrast, our input representations consist of edge-images concatenated with 2D joint heatmaps. To generate edge-images, we first create synthetic RGB images by rendering textured SMPL meshes. For each training mesh, clothing textures are randomly chosen from \cite{varol17_surreal, bhatnagar2019mgn}. The textured SMPL mesh is rendered onto a background image (randomly chosen from LSUN \cite{yu15lsun}), using randomly-sampled lighting and camera parameters. Canny edge detection \cite{canny1986edge} is used to compute edge-images from the synthetic RGB images. We show in Section \ref{subsec:ablation} that, despite the lack of photorealism in the synthetic RGB images, edge-filtering bridges the synthetic-to-real domain gap at test-time - and performs better than either silhouette-based or synthetic-RGB-based training inputs in our experiments. Examples of synthetic training samples are given in the supplementary material.

\begin{table*}[t]
\centering
\small
\renewcommand{\tabcolsep}{2.2pt}
\begin{tabular}{c c c c c c c c c c c c c} 
\hline
\textbf{Input Type} & \textbf{Architecture} & \textbf{2D Samples Loss} & \multicolumn{3}{c}{\textbf{Synthetic Test Data}} & \multicolumn{2}{c}{\textbf{SSP-3D}} & \textbf{3DPW}\\
& & & MPJPE-SC & PVE-T-SC & 2D Joint Err. & PVE-T-SC & 2D Joint Err. & MPJPE-SC\\ 
& & & & &  Mode/Samples & & Mode/Samples &\\ [0.5ex] 
\hline
Silh. + J2DHmap & Independent & No & 84.9 & 12.8 & 7.2 / 11.6 & 14.3 & 6.0 / 11.9 & 93.0 &\\
RGB + J2DHmap & Independent & No & 79.9 & \textbf{11.3} & 7.1 / 11.7 & 14.0 & 5.9 / 12.0 & 92.8\\
Edge + J2DHmap & Independent & No & 85.8 & 12.9 & 7.5 / 12.0 & 13.7 & 5.9 / 11.8 & 88.4\\
\hline
Edge + J2DHmap & Independent & Yes & 86.3 & 13.2 & 7.6 / 8.9 & 13.9 & 6.2 / 9.6 & 91.3\\
Edge + J2DHmap & Hierarchical & No & 84.4 & 12.8 & 7.3 / 10.4 & \textbf{13.6} & 5.3 / 11.2 & 87.7\\
Edge + J2DHmap & Hierarchical & Yes & \textbf{79.1} & 12.6 & \textbf{6.7} / \textbf{6.9} & \textbf{13.6} & \textbf{4.8} / \textbf{6.9} & \textbf{84.7}\\
\hline
\end{tabular}
\vspace{0.2cm}
\caption{Experiments investigating different input representations, hierarchical versus independent distribution prediction networks and the 2D samples reprojection loss, evaluated in terms of shape and pose prediction metrics on synthetic data, SSP-3D \cite{STRAPS2020BMVC} and 3DPW \cite{vonMarcard2018}.}
\label{table:ablation}
\vspace{-0.4cm}
\end{table*}

\noindent \textbf{Training details.} We use Adam \cite{kingma2014adam} with a learning rate of 0.0001, batch size of 80 and train for 150 epochs. For stability, the 2D joints reprojection loss is only applied on the mode pose and shape (projected to 2D) in the first 50 epochs and not on the samples, which are supervised in the next 100 epochs. To boost 3D pose metrics, an MSE loss on the mode 3D joint locations is applied in the final 50 epochs.

\noindent \textbf{Evaluation datasets.} 3DPW \cite{vonMarcard2018} is used to evaluate 3D pose prediction accuracy. We report mean-per-joint-position-error after scale correction (MPJPE-SC) \cite{STRAPS2020BMVC} and after Procrustes analysis (MPJPE-PA), both in mm. Both metrics are computed using the mode 3D joint coordinates of the predicted shape and pose distributions.

SSP-3D is primarily used to evaluate 3D body shape prediction accuracy, using per-vertex Euclidean error in a T-pose after scale-correction (PVE-T-SC) \cite{STRAPS2020BMVC} in mm, computed with the mode 3D body shape from the predicted shape distribution. We also evaluate 2D joint prediction error (2D Joint Err. Mode/Samples) in pixels, computed using both the mode 3D body and 10 3D bodies randomly sampled from the predicted shape and pose distributions, projected onto the image plane using the camera prediction. 2D joint error is evaluated on \textit{visible} target 2D joints only.

Finally, we use a synthetic test dataset for our ablation studies investigating different input representations. It consists of 1000 synthetic input-label pairs, generated in the same way as the synthetic training data, with poses sampled from the test set of Human3.6M. \cite{h36m_pami}.


\section{Experimental Results}
\label{sec:experiments}

This section investigates different input representations and the benefits of the 2D joints samples loss, compares independent and hierarchical distribution predictors and benchmarks our method against the state-of-the-art.

\subsection{Ablation studies}
\label{subsec:ablation}

\noindent \textbf{Input proxy representation.} Rows 1-3 in Table \ref{table:ablation} compare different choices of input proxy representation: binary silhouettes, RGB images and edge-filtered images (each additionally concatenated with 2D joint heatmaps). The independent network architecture is used for all three input types. To investigate the synthetic-to-real domain gap, metrics are presented for synthetic test data, as well as real test images from SSP-3D and 3DPW. For the latter, silhouette segmentation is carried out with DensePose \cite{Guler2018DensePose}. Using RGB-based input representations (row 2) results in the best 3D shape and pose metrics on synthetic data, which is reasonable since RGB contains more information than both silhouettes and edge-filtered images. However, metrics are significantly worse on real datasets, suggesting that the network has over-fitted to unrealistic artefacts present in low-fidelity (i.e. computationally cheap) synthetic RGB images. Silhouette-based input representations (row 1) also demonstrate a deterioration of 3D metrics on real test data compared to synthetic data, since they are heavily reliant upon accurate silhouettes, which are difficult to robustly segment in test images containing challenging poses or severe occlusions. Inaccurate silhouette segmentations critically impair the network's ability to predict 3D body pose and shape. In contrast, edge-filtering is a simpler and more robust operation than segmentation, but is still able to retain important shape information from the RGB image. Thus, edge-images (concatenated with 2D joint heatmaps) can better bridge the synthetic-to-real domain gap, resulting in improved metrics on real test inputs (row 3).

\noindent \textbf{Hierarchical architecture and reprojection loss on 2D joints samples.} Figure \ref{fig:ablation} and rows 3-6 in Table \ref{table:ablation} compare the independent and hierarchical distribution prediction architectures ($f_\text{indep}$ and $f_\text{hier}$) presented in Section \ref{subsec:distribution_predictor}, both with and without the reprojection loss on sampled 2D joints ($\mathcal{L}_\text{2D Samples}$) from Section \ref{subsec:losses}. When $\mathcal{L}_\text{2D Samples}$ is not applied, the shape and pose distributions predicted by both the independent and hierarchical network architectures do not consistently match the the input image, as evidenced by the significant gap between the visible 2D joint error computed using the distributions' modes versus samples drawn from the distributions (in rows 3 and 5 of Table \ref{table:ablation}) on both synthetic test data and SSP-3D \cite{STRAPS2020BMVC}. This implies that the predicted distributions are overly uncertain about parts of the subject's body that are visible and unambiguous in the input image. The visualisations corresponding to the hierarchical architecture trained without $\mathcal{L}_\text{2D Samples}$ in Figure \ref{fig:ablation} (centre) further demonstrate that the predicted samples often do not match the input image, particularly at the extreme ends of the body. This results in significant undesirable per-vertex uncertainty over unambiguous body parts.

Applying $\mathcal{L}_\text{2D Samples}$ to the independent network $f_\text{indep}$ partially alleviates the mismatch between inputs and predicted samples, as shown by Figure \ref{fig:ablation} (right) and row 4 in Table \ref{table:ablation}, where the mode versus sample 2D joint error gap has reduced. However, training with $\mathcal{L}_\text{2D Samples}$ deteriorates the independent architecture's mode pose prediction metrics (MPJPE-SC and 2D Joint Err. Mode in row 3 vs 4 of Table  \ref{table:ablation}) on both synthetic and real test data. This is because $f_\text{indep}$ naively models each joint's relative rotation independently of its parents' rotations (Equation \ref{eqn:indep_mf_joints}); however, to predict realistic human pose samples that match the visible input, each joint’s rotation distribution \textit{must} be informed by its parents. $\mathcal{L}_\text{2D Samples}$ attempts to force predicted samples to match the input despite this logical inconsistency, which causes a trade-off between mode and sample pose prediction metrics, particularly worsening MPJPE-SC.

In contrast, applying $\mathcal{L}_\text{2D Samples}$ to the hierarchical network $f_\text{hier}$ improves metrics corresponding to both mode and sample predictions, as shown by row 6 in Table \ref{table:ablation}. Now, each SMPL joint's relative rotation distribution is conditioned on all its parents' distributions (Equation \ref{eqn:hier_mf_joints}). Thus, $\mathcal{L}_\text{2D Samples}$ and $\mathcal{L}_\text{NLL}$ work in conjunction in enabling predicted hierarchical distributions (and samples) to match the visible input, while yielding improved 3D metrics. Figure \ref{fig:ablation} (left) exhibits such visually-consistent samples and demonstrates greater prediction uncertainty for ambiguous parts. Note that uncertainty can arise even without occlusion in a monocular setting, e.g. due to depth ambiguities \cite{Sminchisescu2001covsampling, Sminchisescu2003kinematicjump} as shown by the left arm samples in the last row of Figure \ref{fig:ablation}. Further visual results are in the supplementary material.

\begin{table}[t]
\centering
\small
\renewcommand{\tabcolsep}{2.2pt}
\begin{tabular}{l c c c} 
 \hline
 \textbf{Method} & \multicolumn{3}{c}{\textbf{3DPW}}\\
 & MPJPE & MPJPE-SC & MPJPE-PA\\ [0.5ex] 
 \hline
 HMR \cite{hmrKanazawa17} & 130.0 & 102.8 & 76.7\\
 GraphCMR \cite{kolotouros2019cmr} & 119.9 & 102.0 & 70.2\\
 SPIN \cite{kolotouros2019spin} & 96.9 & 89.4 & 59.0\\
 Pose2Mesh \cite{Choi_2020_ECCV_Pose2Mesh} & 89.2 & - & 58.9\\
 I2L-MeshNet \cite{Moon_2020_ECCV_I2L-MeshNet} & 93.2 & 77.5 & 57.7\\
 Biggs \etal \cite{biggs2020multibodies} & 93.8 & - & 59.9\\
 DaNet \cite{zhang2019danet} & 85.5 & 76.4 & 54.8\\
 HybrIK \cite{li2020hybrik} & \textbf{80.0} & - & \textbf{48.8}\\
 \hline
 HMR (unpaired) \cite{hmrKanazawa17} & - & 126.3 & 92.0\\
 Kundu \etal \cite{kundu_human_mesh} & 153.4 & - & 89.8\\
 STRAPS \cite{STRAPS2020BMVC} & - & 99.0 & 66.8\\
 Sengupta \etal \cite{sengupta2021probabilisticposeshape} & - & 90.9 & 61.0\\
 Ours w. Detectron2 \cite{wu2019detectron2} & 96.2 & 84.7 & 59.2\\
 Ours w. HRNet-W48 \cite{sun2019hrnet} & \textbf{84.9} & \textbf{73.0} & \textbf{53.6}\\
 \hline
\end{tabular}
\vspace{0.2cm}
\caption{Comparison with SOTA in terms of MPJPE, MPJPE-SC and MPJPE-PA (all mm) on 3DPW \cite{vonMarcard2018}. Methods in the top half require training images paired with 3D ground-truth, methods in the bottom half do not. For our method, we present metrics using both Detectron2 \cite{wu2019detectron2} and HRNet \cite{sun2019hrnet} as 2D joint detectors for proxy representation computation, to enable a fair comparison with past methods \cite{STRAPS2020BMVC, sengupta2021probabilisticposeshape} that used Detectron2.}
\label{table:3dpw_sota_comparison}
\vspace{-0.3cm}
\end{table}

\begin{table}[t!]
\centering
\small
\begin{tabular}{c l c c} 
 \hline
 \multirow{2}{0.23\linewidth}{\textbf{Max. input set size}} & \textbf{Method} & \textbf{SSP-3D}\\
 & & PVE-T-SC\\ [0.5ex] 
 \hline
 & HMR \cite{hmrKanazawa17} & 22.9\\
 & GraphCMR \cite{kolotouros2019cmr} & 19.5\\
  \large 1 & SPIN \cite{kolotouros2019spin} & 22.2\\
 & DaNet \cite{zhang2019danet} & 22.1\\
 & STRAPS \cite{STRAPS2020BMVC} & 15.9\\
 & Sengupta \etal \cite{sengupta2021probabilisticposeshape} & 15.2\\
 & Ours & \textbf{13.6}\\
 \hline
 & HMR \cite{hmrKanazawa17} + Mean & 22.9\\
 & GraphCMR \cite{kolotouros2019cmr} + Mean & 19.3\\
 & SPIN \cite{kolotouros2019spin} + Mean & 21.9\\
 \large 5 & DaNet \cite{zhang2019danet} + Mean & 22.1\\
 & STRAPS \cite{STRAPS2020BMVC} + Mean & 14.4 \\
 & Sengupta \etal \cite{sengupta2021probabilisticposeshape} + Mean & 13.6\\
 & Sengupta \etal \cite{sengupta2021probabilisticposeshape} + Prob. Comb. & 13.3\\
 & Ours + Mean & 12.2\\
 & Ours + Prob. Comb. & \textbf{12.0}\\
 \hline
\end{tabular}
\vspace{0.1cm}
\caption{Comparison with SOTA in terms of PVE-T-SC (mm) on SSP-3D \cite{STRAPS2020BMVC}. Top half: single-input, bottom half: multi-input. Prob. Comb. denotes the multi-input probabilistic combination approach proposed in \cite{sengupta2021probabilisticposeshape}.}
\label{table:ssp3d_sota_comparison}
\vspace{-0.1cm}
\end{table}

\begin{figure}[t!]
    \centering
    \includegraphics[width=\columnwidth]{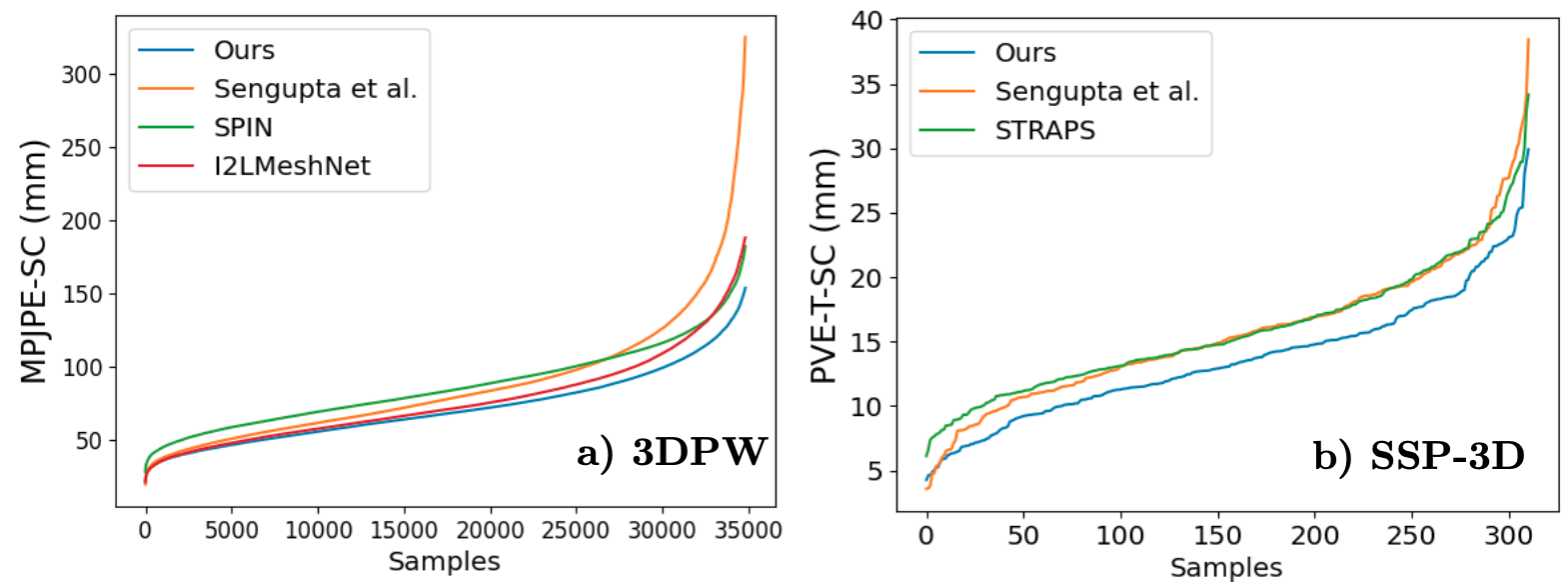}
    \caption{Comparison with SOTA using sorted per-sample distributions of a) MPJPE-SC on 3DPW and b) PVE-T-SC on SSP-3D.}
    \label{fig:error_dist}
    \vspace{-0.3cm}
\end{figure}

\subsection{Comparison with the state-of-the-art}

\begin{figure*}[b!]
    \centering
    \includegraphics[width=\textwidth]{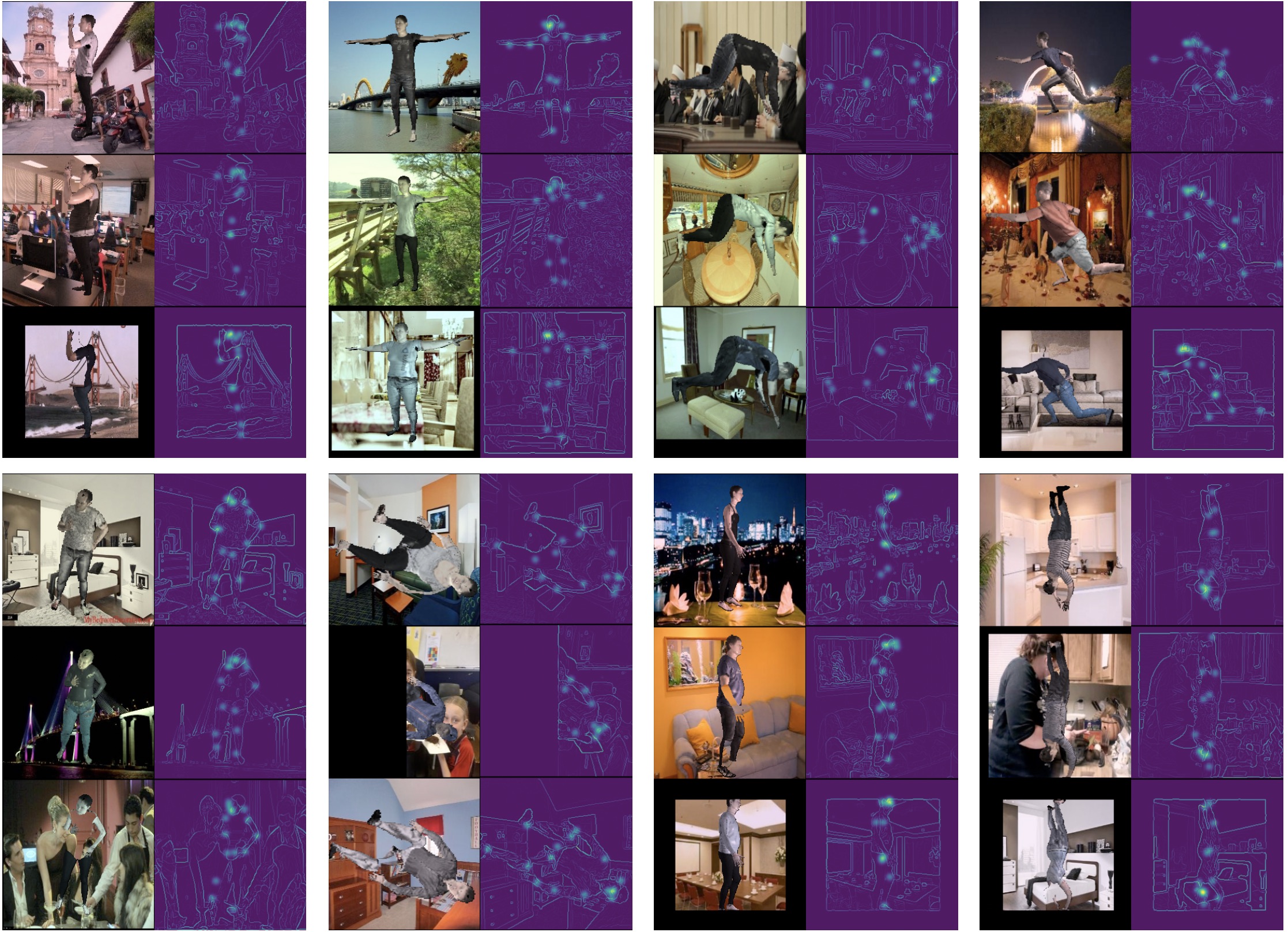}
    \vspace{0.1cm}
    \caption{Examples of synthetic training and validation data rendered on-the-fly during model training. Synthetic RGB images are converted into edge-filtered images and 2D joint heatmaps, which act as the input to the distribution prediction network presented in the main manuscript. The synthetic RGB images are computationally-cheap and far from photorealistic. However, edge detection \cite{canny1986edge} is able to significantly bridge the synthetic-to-real domain gap, as can be seen by comparing the synthetic edge-images with real edge-images in Figures \ref{fig:supmat_3dpw} and \ref{fig:supmat_ssp3d} of the supplementary material.}
    \label{fig:supmat_training_data}
\end{figure*}

\noindent \textbf{Shape prediction.} Table \ref{table:ssp3d_sota_comparison} evaluates 3D body shape metrics on SSP-3D \cite{STRAPS2020BMVC} for single image inputs and multi-image input sets, which we evaluate using both mean and probabilistic combination methods from \cite{sengupta2021probabilisticposeshape}. Our network surpasses the state-of-the-art \cite{sengupta2021probabilisticposeshape}, mainly due to our use of an edge-based proxy representation, instead of the silhouette-based representations used in \cite{STRAPS2020BMVC} and \cite{sengupta2021probabilisticposeshape}. These methods rely on accurate human silhouettes, which may be difficult to compute at test-time, as discussed in Section \ref{subsec:ablation}, while our method does not have such dependencies. However, our method may result in erroneous shape predictions when the subject is wearing loose clothing which obscures body shape, in which case the shape prediction over-estimates the subject's true proportions (see rows 1-2 in Figure \ref{fig:ablation}).

\noindent \textbf{Pose prediction.} Table \ref{table:3dpw_sota_comparison} evaluates 3D pose metrics on 3DPW \cite{vonMarcard2018}. Our method is competitive with the state-of-the-art and surpasses other methods that do not require 3D-labelled training images \cite{STRAPS2020BMVC, sengupta2021probabilisticposeshape, kundu_human_mesh, hmrKanazawa17}. Figure \ref{fig:error_dist}(a) shows that our method performs well for most test examples in 3DPW, even matching pose-focused approaches that do not attempt to accurately predict diverse body shapes \cite{Moon_2020_ECCV_I2L-MeshNet, kolotouros2019spin}. However, some images in 3DPW contain significant occlusion, which can lead to noisy 2D joint heatmaps in the proxy representations, resulting in poor 3D pose metrics as shown by the right end of the curve in Figure \ref{fig:error_dist}(a).

Further quantitative comparison with other shape and pose distribution/multi-hypothesis prediction approaches is given in the supplementary material.

\section{Conclusion}
In this paper, we have proposed a probabilistic approach to the ill-posed problem of monocular 3D human shape and pose estimation, motivated by the fact that multiple 3D bodies may explain a given 2D image. Our method predicts a novel hierarchical matrix-Fisher distribution over relative 3D joint rotations and a Gaussian distribution over SMPL \cite{SMPL:2015} shape parameters, from which we can sample any number of plausible 3D reconstructions. To ensure that the predicted distributions match the input image, we have implemented a differentiable rejection sampler to impose a loss between predicted 2D joint samples and ground-truth 2D joint coordinates. Our method is competitive with the state-of-the-art in terms of pose metrics on 3DPW, while surpassing the state-of-the-art for shape accuracy on SSP-3D.

\noindent \textbf{Acknowledgements.} We thank Dr. Yu Chen (Metail), Mr. Jim Downing (Metail), Dr. David Bruner (SizeStream) and Dr. Delman Lee (TAL Apparel) for providing body shape evaluation data and supporting this research.

\twocolumn[
\large
\centering
\textbf{Supplementary Material: Hierarchical Kinematic Probability Distributions for 3D Human Shape and Pose Estimation from Images in the Wild\\}
\vspace{0.2in}
]

\noindent Section \ref{sec:supmat_implementation_details} in this supplementary material contains implementation details, particularly regarding synthetic training data generation and per-vertex uncertainty visualisation. Section \ref{sec:supmat_qualitative_results} discusses qualitative results on the SSP-3D \cite{STRAPS2020BMVC} and 3DPW \cite{vonMarcard2018} datasets, and compares distribution predictions on images with versus without artificial occlusions. Table \ref{table:sup_mat_multihypothesis_eval} compares several recent multi-hypothesis 3D human shape and pose estimation approaches.

\section{Implementation Details}
\label{sec:supmat_implementation_details}

\subsection{Synthetic Training Data}

Our shape and pose distribution prediction neural networks are trained using synthetic training data, consisting of edge-and-joint-heatmap inputs paired with ground truth SMPL \cite{SMPL:2015} shape and pose parameters. Inputs are rendered on-the-fly during model training using randomly sampled camera extrinsics, lighting, backgrounds and clothing textures. Examples of synthetic training and validation data are given in Figure \ref{fig:supmat_training_data}. Note how each body pose may be paired with a different body shape, clothing, camera and background, as well as occlusion and noise augmentations. Thus, we are able to render highly diverse training data on-the-fly during training, enabling the network to see a new pose/shape/clothing/camera/background combination in each training iteration.

Our synthetic RGB images (Figure \ref{fig:supmat_training_data}) are computationally cheap but clearly far from photorealistic, resulting in a large synthetic-to-real domain gap. However, simple edge detection \cite{canny1986edge} is able to significantly reduce this gap \cite{Charles2020realtimesscreen}, motivating the use of edge-filtered images as part of our input proxy representation. We found that noisy edge detections (as seen in  Figure \ref{fig:supmat_training_data}) retained sufficient visual shape and pose information, and efforts to produce clean edge-images (e.g. hysteresis-based edge tracking or further hyperparameter tuning) did not improve performance.

The required body shape, pose, clothing and backgrounds are obtained as follows. For training, ground-truth SMPL 3D joint rotation matrices are sampled from the training splits of 3DPW \cite{vonMarcard2018} and UP-3D \cite{Lassner:UP:2017}, as well as Human3.6M \cite{h36m_pami} subjects 1, 5, 6, 7 and 8, giving a total of 91106 training poses. Validation poses are sampled from the 3DPW/UP-3D validation splits and Human3.6M subjects 9 and 11, resulting in 33347 validation poses. SMPL body shape parameters are randomly sampled from $\mathcal{N}(\beta_i; 0, 1.25^2)$ for $i=1, ..., 10$ \cite{STRAPS2020BMVC}. RGB clothing textures for the SMPL body mesh are selected from SURREAL \cite{varol17_surreal} and MultiGarmentNet \cite{bhatnagar2019mgn}, resulting in 917 training textures and 108 validation textures. Backgrounds are obtained from LSUN \cite{yu15lsun}, which contains a collection of diverse indoor and outdoor scenes. We sample from 397582 different training backgrounds and 3000 different validation backgrounds. Note that background training images may contain other humans, which is intentional and essential for robustness against test images with multiple people. The network learns to focus on the person corresponding to the input joint heatmaps and ignore persons in the background.

Textured SMPL meshes are rendered with Pytorch3D \cite{ravi2020pytorch3d}, using a perspective camera model and Phong shading. Camera and lighting parameters are randomly sampled, with sampling hyperparameters given in Table \ref{table:sup_mat_generate_hypparams}. Generated images are cropped around the rendered body using a square bounding box, where the bounding box size is randomly scaled by a factor in range (0.8, 1.2).

To further bridge the gap synthetic-to-real gap, we implement random occlusion, body part removal, 2D joint removal and 2D joint noise augmentations during training. Hyperparameters associated with data augmentations are given in Table \ref{table:sup_mat_augment_hypparams}.

\begin{table}[t]
\centering
\small
\begin{tabular}{l c}
    \hline
    \noalign{\smallskip} 
    \textbf{Hyperparameter} & \textbf{Value}\\
    \noalign{\smallskip}
    \hline
    \noalign{\smallskip}
    Shape parameter sampling mean & 0 \\
    Shape parameter sampling std. & 1.25 \\
    Cam. translation sampling mean & (0, -0.2, 2.5) m\\
    Cam. translation sampling var. & (0.05, 0.05, 0.25) m\\
    Cam. focal length & 300.0\\
    Lighting ambient intensity range & [0.4, 0.8]\\
    Lighting diffuse intensity range & [0.4, 0.8]\\
    Lighting specular intensity range & [0.0, 0.5]\\
    Bounding box scale factor range & [0.8, 1.2]\\
    Proxy representation dimensions & $256 \times 256$ pixels\\
    \noalign{\smallskip}
    \hline
    \noalign{\smallskip}
    \noalign{\smallskip}
    \end{tabular}
\caption{List of hyperparameter values associated with synthetic training data generation.}
\label{table:sup_mat_generate_hypparams}
\end{table}

\subsection{Visualisation of Per-Vertex Uncertainty}
\begin{figure*}[t!]
    \centering
    \includegraphics[width=\textwidth]{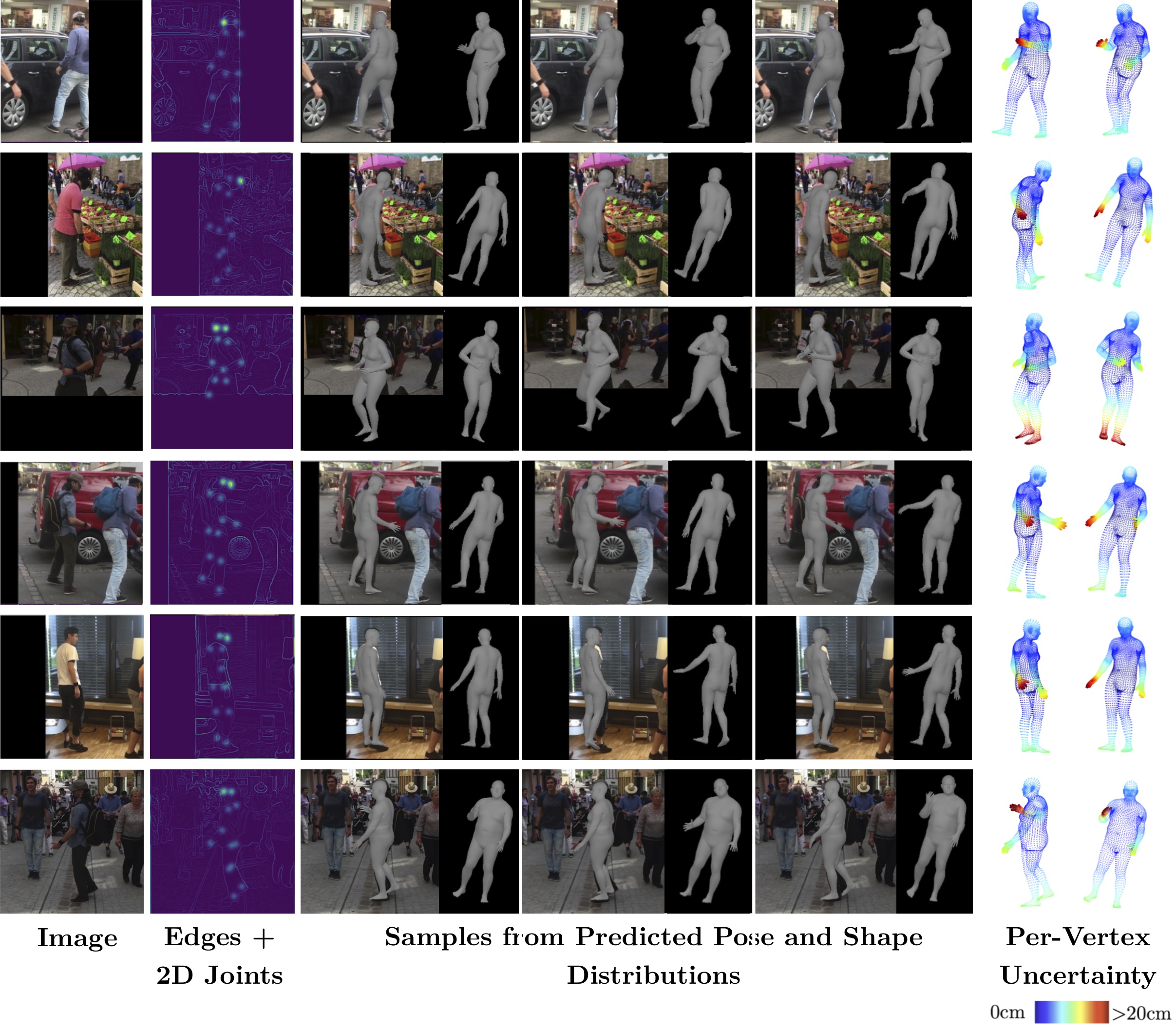}
    \caption{3D reconstruction samples and per-vertex uncertainties corresponding to shape and relative 3D joint rotation distributions predicted from 3DPW images\cite{vonMarcard2018}. The selected images exhibit self-occlusion and out-of-frame body parts, which result in greater 3D location uncertainty for vertices belonging to ambiguous parts.}
    \label{fig:supmat_3dpw}
\end{figure*}

Figures \ref{fig:supmat_3dpw}, \ref{fig:supmat_ssp3d} and \ref{fig:supmat_ssp3d_occlude_compare} in this supplementary material, as well as several figures in the main manuscript, visualise per-vertex 3D location uncertainties corresponding to the predicted shape and 3D joint rotation distributions. These are computed by i) sampling 100 shape parameter vectors and relative 3D joint rotations (for the entire kinematic tree) from the predicted distributions, ii) passing each of these samples through the SMPL function \cite{SMPL:2015} to get the corresponding vertex meshes, iii) computing the mean location of each vertex over all the samples and iv) determining the average Euclidean distance from the sample mean for each vertex over all the samples, which is ultimately visualised in the vertex scatter plots as a measure of per-vertex 3D location uncertainty.

\begin{table*}[h!]
\centering
\renewcommand{\tabcolsep}{6.0pt}
\begin{tabular}{l | c c c c | c c c c | c c c c } 
 \hline
 \textbf{Method} & \multicolumn{8}{c}{\textbf{3DPW}} & \multicolumn{4}{|c}{\textbf{SSP-3D}}\\
 & \multicolumn{4}{c}{MPJPE} & \multicolumn{4}{|c}{MPJPE-PA} & \multicolumn{4}{|c}{PVE-T-SC}\\ 
Number of Samples: & 1 & 5 & 10 & 25 & 1 & 5 & 10 & 25 & 1 & 5 & 10 & 25\\[0.5ex]
 \hline
 Biggs \etal \cite{biggs2020multibodies} & 93.8 & 82.2 & 79.4 & 75.8 & 59.9 & 57.1 & 56.6 & 55.6 & - & - & - & -\\ 
 Sengupta \etal \cite{sengupta2021probabilisticposeshape} & 97.1 & 95.8 & 93.1 & 89.7 & 61.1 & 59.4 & 58.2 & 56.5 & 15.2 & 14.8 & 13.6 & 11.9\\
 ProHMR \cite{kolotouros2021prohmr} & - & - & - & - & 59.8 & 56.5 & 54.6 & 52.4 & - & - & - & -\\
 Ours (Independent) w. HRNet \cite{sun2019hrnet} & 88.3 & 85.0 & 82.6 & 78.5 & 56.6 & 54.5 & 52.8 & 50.2 & 13.9 & 12.9 & 12.0 & 10.3\\
 Ours (Hierarchical) w. HRNet \cite{sun2019hrnet}  & \textbf{84.9} & \textbf{81.6} & \textbf{79.0} & \textbf{75.1} & \textbf{53.6} & \textbf{51.4} & \textbf{49.6} & \textbf{47.0} & \textbf{13.6} & \textbf{12.3} & \textbf{11.3} & \textbf{9.8} \\
 \hline
\end{tabular}
\vspace{0.2cm}
\caption{Comparison with other 3D human shape and pose distribution/multi-hypothesis estimation methods. Following Biggs \etal \cite{biggs2020multibodies}, we report body shape (PVE-T-SC) and pose (MPJPE and MPJPE-PA) metrics computed using the minimum error sample out of a set of $n$ predicted samples for each test image, for $n \in \{1, 5, 10 ,25\}$. This is motivated by the fact that the single ground-truth 3D body only represents \textit{one plausible 3D solution out of many} (for ambiguous images), which may not be the same as the mode of our predicted shape and pose distribution. The improvement in 3D shape and pose metrics with increasing number of samples shows that our predicted distribution is able to model the 3D ground-truth as a possible sample.}
\label{table:sup_mat_multihypothesis_eval}
\end{table*}


\begin{table}[t]
\centering
\small
\begin{tabular}{l l c}
    \hline
    \noalign{\smallskip} 
    \textbf{Augmentation} & \textbf{Hyperparameter} & \textbf{Value}\\
    \noalign{\smallskip}
    \hline
    \noalign{\smallskip}
    Body part occlusion & Occlusion probability & 0.1 \\
    2D joints L/R swap & Swap probability & 0.1\\
    Half-image occlusion & Occlusion probability & 0.05\\
    2D joints removal & Removal probability & 0.1\\
    2D joints noise & Noise range & [-8, 8] pixels\\
    Occlusion box & Probability, Size & 0.5, 48 pixels \\
    \noalign{\smallskip}
    \hline
    \noalign{\smallskip}
    \noalign{\smallskip}
    \end{tabular}
\caption{List of synthetic training data augmentations and their associated hyperparameter values. Body part occlusion uses the 24 DensePose \cite{Guler2018DensePose} parts. Joint L/R swap is done for shoulders, elbows, wrists, hips, knees, ankles.}
\label{table:sup_mat_augment_hypparams}
\end{table}

\section{Qualitative Results}
\label{sec:supmat_qualitative_results}

\begin{figure*}[t!]
    \centering
    \includegraphics[width=\textwidth]{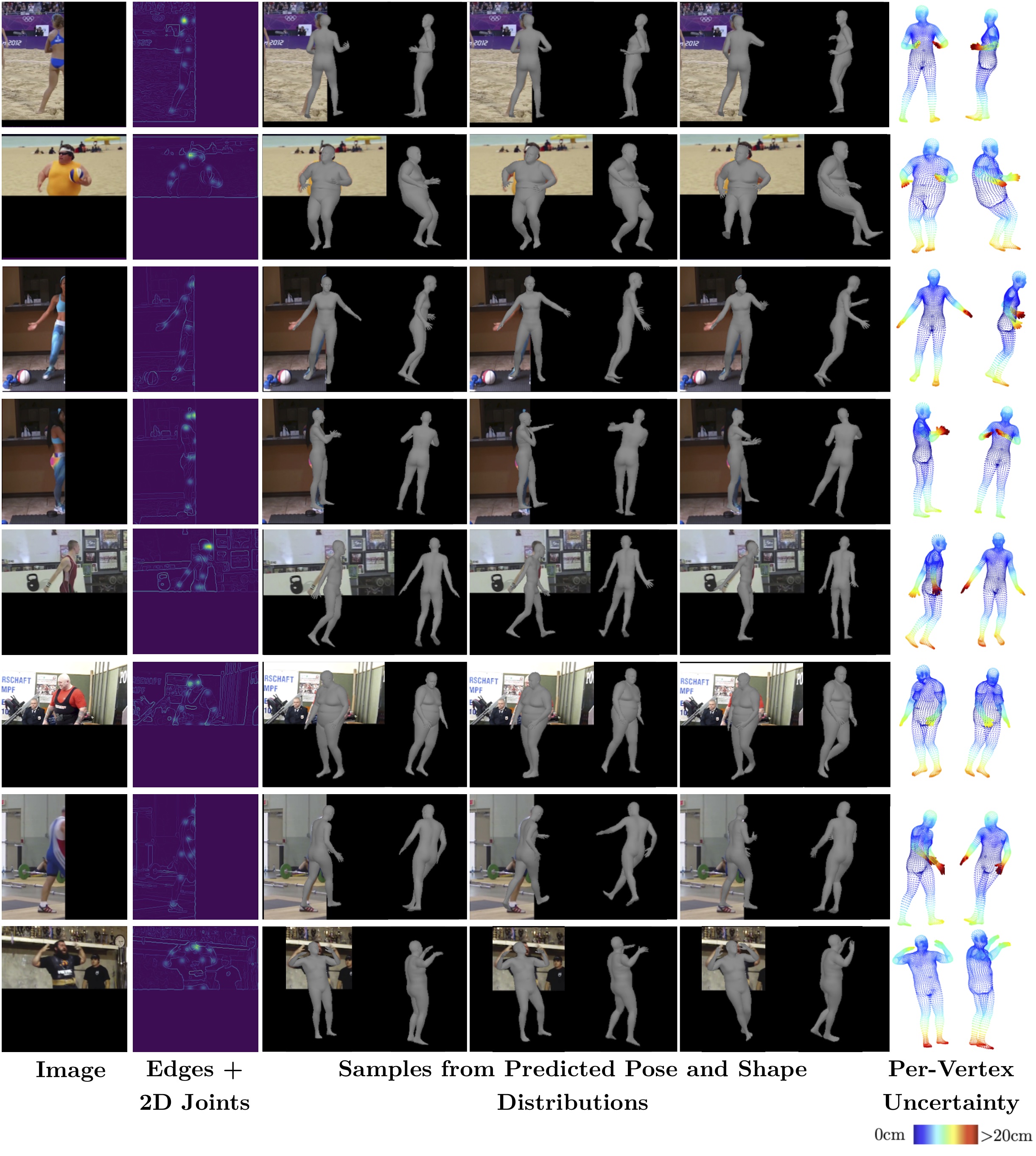}
    \caption{3D reconstruction samples and per-vertex uncertainties corresponding to shape and relative 3D joint rotation distributions predicted from SSP-3D images\cite{STRAPS2020BMVC}. The images are artificially occluded, resulting in greater 3D location uncertainty for vertices belonging to ambiguous parts.}
    \label{fig:supmat_ssp3d}
\end{figure*}

Figure \ref{fig:supmat_ssp3d} presents results on artificially occluded images from SSP-3D \cite{STRAPS2020BMVC}. In particular, note that i) occluded/invisible body parts result in increased 3D location uncertainty for corresponding vertices and ii) 3D body samples from the predicted distributions match the visible body parts in the 2D image, while invisible body part samples are more diverse. However, occluded sample diversity is still somewhat limited and samples tend to be clustered around the mode predictions, which is a weakness of our method. This may be alleviated by predicting multi-modal distributions over 3D shape and pose in future work. Figure \ref{fig:supmat_ssp3d} also illustrates our method's ability to predict a range of body shapes, owing to the synthetic training framework used.

Figure \ref{fig:supmat_3dpw} presents results on the test split of 3DPW \cite{vonMarcard2018}. Again, note the increased uncertainty and sample diversity for occluded and out-of-frame body parts, and the reprojection consistency between predicted samples and the visible bodies in the images. Results on 3DPW highlight another key challenge for future work: when faced with baggy/loose clothing, our method tends to over-estimate the subject's body proportions. This is because our synthetic training data does not model the shape of clothing on the human body surface, but only its texture. Future work could focus on using synthetic \textit{clothed} humans for training.

\begin{figure*}[t!]
    \centering
    \includegraphics[width=\textwidth]{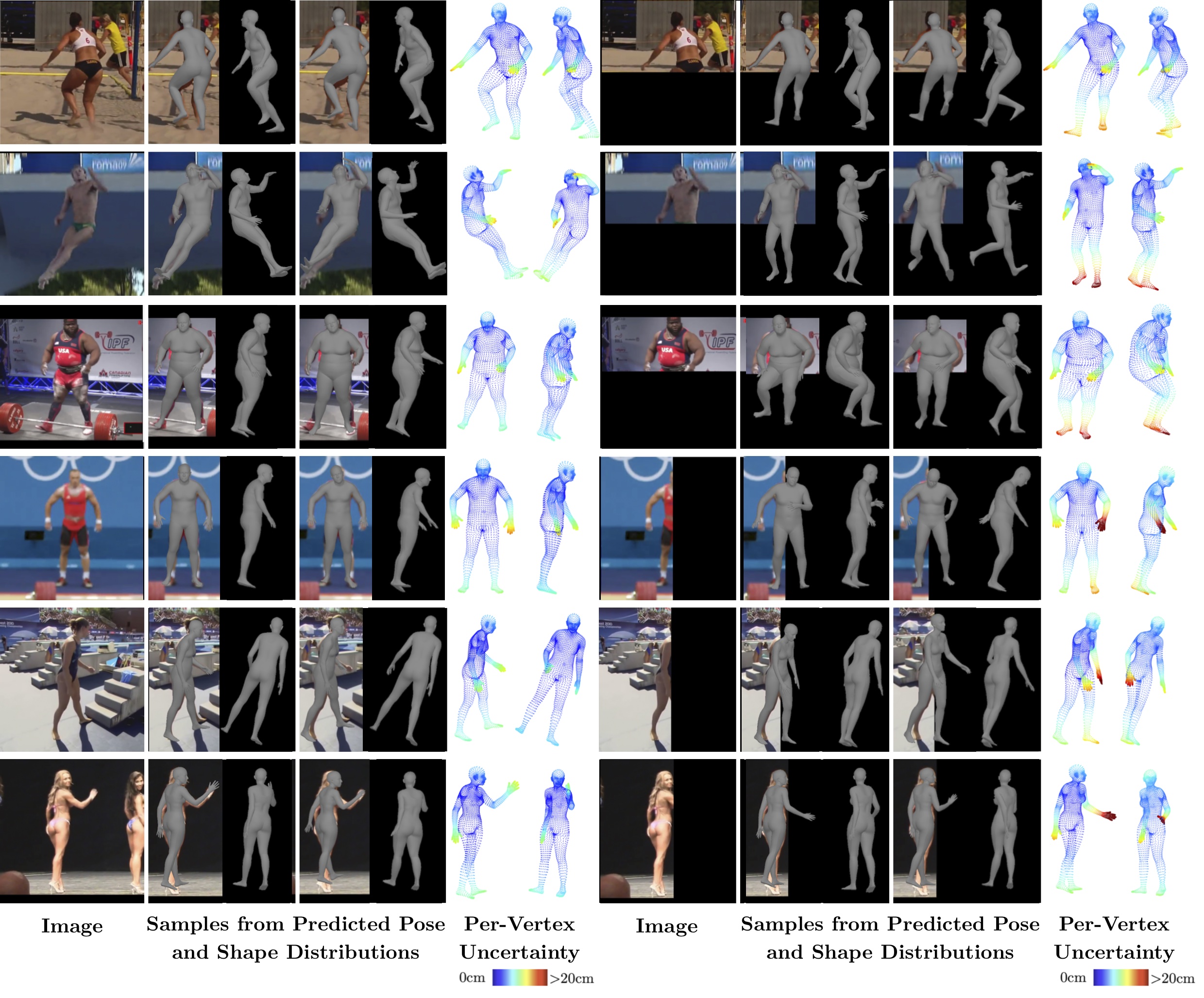}
    \caption{Comparison between 3D samples and per-vertex uncertainties obtained using artificially occluded versus non-occluded input images from SSP-3D \cite{STRAPS2020BMVC}. Ambiguous parts have greater prediction uncertainty.}
    \label{fig:supmat_ssp3d_occlude_compare}
\end{figure*}

Figure \ref{fig:supmat_ssp3d_occlude_compare} compares shape and pose distribution predictions on images from SSP-3D with versus without artificial occlusions, further corroborating that ambiguous parts result in greater uncertainty and more diverse 3D samples. However, it is again apparent that sample diversity for highly ambiguous parts is more limited than expected, as samples tend to be closely clustered around the mode prediction. 

Note that uncertainty does not only arise from occlusion - depth ambiguities are prevalent when estimating 3D pose from a monocular 2D image \cite{Sminchisescu2001covsampling, Sminchisescu2003kinematicjump}. This is demonstrated in the non-occluded images in Figure \ref{fig:supmat_ssp3d_occlude_compare} (left), by the left arm samples in rows 1 and 5 and the right arm in row 4. 

\clearpage
\clearpage
{\small
\bibliographystyle{ieee_fullname}
\bibliography{egbib}

\begin{thebibliography}{10}\itemsep=-1pt

\bibitem{Anguelov05scape:shape}
Dragomir Anguelov, Praveen Srinivasan, Daphne Koller, Sebastian Thrun, Jim
  Rodgers, and James Davis.
\newblock {SCAPE}: Shape completion and animation of people.
\newblock In {\em ACM Transactions on Graphics (TOG) - Proceedings of
  SIGGRAPH}, volume~24, pages 408--416, 2005.

\bibitem{bhatnagar2019mgn}
Bharat~Lal Bhatnagar, Garvita Tiwari, Christian Theobalt, and Gerard Pons-Moll.
\newblock Multi-garment net: Learning to dress {3D} people from images.
\newblock In {\em Proceedings of the {IEEE} International Conference on
  Computer Vision ({ICCV})}, Oct 2019.

\bibitem{biggs2020multibodies}
Benjamin Biggs, S{\'{e}}bastien Erhardt, Hanbyul Joo, Benjamin Graham, Andrea
  Vedaldi, and David Novotny.
\newblock {3D} multibodies: Fitting sets of plausible {3D} models to ambiguous
  image data.
\newblock In {\em NeurIPS}, 2020.

\bibitem{Bishop94mixturedensity}
Christopher~M. Bishop.
\newblock Mixture density networks.
\newblock Technical report, 1994.

\bibitem{Bogo:ECCV:2016}
Federica Bogo, Angjoo Kanazawa, Christoph Lassner, Peter Gehler, Javier Romero,
  and Michael~J. Black.
\newblock Keep it {SMPL}: Automatic estimation of {3D} human pose and shape
  from a single image.
\newblock In {\em Proceedings of the European Conference on Computer Vision
  (ECCV)}, Oct. 2016.

\bibitem{canny1986edge}
John~F. Canny.
\newblock A computational approach to edge detection.
\newblock {\em IEEE Transactions on Pattern Analysis and Machine Intelligence
  (PAMI)}, 8(6):679--698, 1986.

\bibitem{Charles2020realtimesscreen}
J. Charles, S. Bucciarelli, and R. Cipolla.
\newblock Real-time screen reading: reducing domain shift for one-shot
  learning.
\newblock In {\em Proceedings of the British Machine Vision Conference (BMVC)},
  2020.

\bibitem{Choi_2020_ECCV_Pose2Mesh}
Hongsuk Choi, Gyeongsik Moon, and Kyoung~Mu Lee.
\newblock Pose2mesh: Graph convolutional network for {3D} human pose and mesh
  recovery from a {2D} human pose.
\newblock In {\em Proceedings of the European Conference on Computer Vision
  (ECCV)}, 2020.

\bibitem{Choo2001tracking}
Kiam Choo and D.J. Fleet.
\newblock People tracking using hybrid monte carlo filtering.
\newblock In {\em Proceedings of the IEEE International Conference on Computer
  Vision (ICCV)}, volume~2, pages 321--328 vol.2, 2001.

\bibitem{Deutscher2000particle}
J. Deutscher, A. Blake, and I. Reid.
\newblock Articulated body motion capture by annealed particle filtering.
\newblock In {\em Proceedings of the IEEE Conference on Computer Vision and
  Pattern Recognition (CVPR)}, 2000.

\bibitem{down1972orientationstatistics}
Thomas~D. Downs.
\newblock {Orientation statistics}.
\newblock {\em Biometrika}, 59(3):665--676, 12 1972.

\bibitem{georgakis2020hkmr}
Georgios Georgakis, Ren Li, Srikrishna Karanam, Terrence Chen, Jana Kosecka,
  and Ziyan Wu.
\newblock Hierarchical kinematic human mesh recovery.
\newblock In {\em Proceedings of the European Conference on Computer Vision
  (ECCV)}, 2020.

\bibitem{Gilitschenski2020}
Igor Gilitschenski, Roshni Sahoo, Wilko Schwarting, Alexander Amini, Sertac
  Karaman, and Daniela Rus.
\newblock Deep orientation uncertainty learning based on a bingham loss.
\newblock In {\em International Conference on Learning Representations}, 2020.

\bibitem{Guler_2019_CVPR_holopose}
Riza~Alp Guler and Iasonas Kokkinos.
\newblock Holopose: Holistic {3D} human reconstruction in-the-wild.
\newblock In {\em Proceedings of the IEEE Conference on Computer Vision and
  Pattern Recognition (CVPR)}, June 2019.

\bibitem{Guler2018DensePose}
Riza~Alp G\"uler, Natalia Neverova, and Iasonas Kokkinos.
\newblock Densepose: Dense human pose estimation in the wild.
\newblock In {\em Proceedings of IEEE Conference on Computer Vision and Pattern
  Recognition (CVPR)}, 2018.

\bibitem{h36m_pami}
Catalin Ionescu, Dragos Papava, Vlad Olaru, and Cristian Sminchisescu.
\newblock {Human3.6M}: Large scale datasets and predictive methods for {3D}
  human sensing in natural environments.
\newblock {\em IEEE Transactions on Pattern Analysis and Machine Intelligence
  (PAMI)}, 36(7):1325--1339, July 2014.

\bibitem{Jahangiri2017ICCVW}
Ehsan Jahangiri and Alan~L. Yuille.
\newblock Generating multiple diverse hypotheses for human {3D} pose consistent
  with {2D} joint detections.
\newblock In {\em IEEE International Conference on Computer Vision (ICCV)
  Workshops (PeopleCap)}, 2017.

\bibitem{Joo_2018_CVPR_total_capture}
Hanbyul Joo, Tomas Simon, and Yaser Sheikh.
\newblock Total capture: A {3D} deformation model for tracking faces, hands,
  and bodies.
\newblock In {\em Proceedings of the IEEE Conference on Computer Vision and
  Pattern Recognition (CVPR)}, June 2018.

\bibitem{hmrKanazawa17}
Angjoo Kanazawa, Michael~J. Black, David~W. Jacobs, and Jitendra Malik.
\newblock End-to-end recovery of human shape and pose.
\newblock In {\em Proceedings of the IEEE Conference on Computer Vision and
  Pattern Recognition (CVPR)}, 2018.

\bibitem{kent2013binghamsampling}
John~T. Kent, Asaad~M. Ganeiber, and Kanti~V. Mardia.
\newblock A new method to simulate the {Bingham} and related distributions in
  directional data analysis with applications, 2013.

\bibitem{khatri1977vonmisesfisher}
C.~G. Khatri and K.~V. Mardia.
\newblock The {Von Mises-Fisher} matrix distribution in orientation statistics.
\newblock {\em Journal of the Royal Statistical Society. Series B
  (Methodological)}, 39(1):95--106, 1977.

\bibitem{kingma2014adam}
Diederik~P. Kingma and Jimmy Ba.
\newblock Adam: A method for stochastic optimization.
\newblock In {\em Proceedings of the International Conference on Learning
  Representations (ICLR)}, 2014.

\bibitem{kingma2014autoencoding}
Diederik~P Kingma and Max Welling.
\newblock Auto-encoding variational bayes, 2014.

\bibitem{kirillov2019pointrend}
Alexander Kirillov, Yuxin Wu, Kaiming He, and Ross Girshick.
\newblock {PointRend}: Image segmentation as rendering.
\newblock In {\em Proceedings of the IEEE Conference on Computer Vision and
  Pattern Recognition (CVPR)}, 2020.

\bibitem{kolotouros2019spin}
Nikos Kolotouros, Georgios Pavlakos, Michael~J Black, and Kostas Daniilidis.
\newblock Learning to reconstruct {3D} human pose and shape via model-fitting
  in the loop.
\newblock In {\em Proceedings of the IEEE International Conference on Computer
  Vision (ICCV)}, 2019.

\bibitem{kolotouros2019cmr}
Nikos Kolotouros, Georgios Pavlakos, and Kostas Daniilidis.
\newblock Convolutional mesh regression for single-image human shape
  reconstruction.
\newblock In {\em Proceedings of the IEEE Conference on Computer Vision and
  Pattern Recognition (CVPR)}, 2019.

\bibitem{kolotouros2021prohmr}
Nikos Kolotouros, Georgios Pavlakos, Dinesh Jayaraman, and Kostas Daniilidis.
\newblock Probabilistic modeling for human mesh recovery.
\newblock In {\em ICCV}, 2021.

\bibitem{kundu_human_mesh}
Jogendra~Nath Kundu, Mugalodi Rakesh, Varun Jampani, Rahul~M Venkatesh, and
  R.~Venkatesh Babu.
\newblock Appearance consensus driven self-supervised human mesh recovery.
\newblock In {\em Proceedings of the European Conference on Computer Vision
  (ECCV)}, 2020.

\bibitem{Lassner:UP:2017}
Christoph Lassner, Javier Romero, Martin Kiefel, Federica Bogo, Michael~J.
  Black, and Peter~V. Gehler.
\newblock {Unite the People}: Closing the loop between {3D} and {2D} human
  representations.
\newblock In {\em Proceedings of the IEEE Conference on Computer Vision and
  Pattern Recognition (CVPR)}, 2017.

\bibitem{lee2018bayesianattitude}
T. {Lee}.
\newblock Bayesian attitude estimation with the matrix fisher distribution on
  so(3).
\newblock {\em IEEE Transactions on Automatic Control}, 63(10):3377--3392,
  2018.

\bibitem{Li_2019_CVPR}
Chen Li and Gim~Hee Lee.
\newblock Generating multiple hypotheses for {3D} human pose estimation with
  mixture density network.
\newblock In {\em The IEEE Conference on Computer Vision and Pattern
  Recognition (CVPR)}, June 2019.

\bibitem{li2020hybrik}
Jiefeng Li, Chao Xu, Zhicun Chen, Siyuan Bian, Lixin Yang, and Cewu Lu.
\newblock Hybrik: A hybrid analytical-neural inverse kinematics solution for 3d
  human pose and shape estimation.
\newblock In {\em CVPR}, 2021.

\bibitem{SMPL:2015}
Matthew Loper, Naureen Mahmood, Javier Romero, Gerard Pons-Moll, and Michael~J.
  Black.
\newblock {SMPL}: A skinned multi-person linear model.
\newblock In {\em ACM Transactions on Graphics (TOG) - Proceedings of ACM
  SIGGRAPH Asia}, volume~34, pages 248:1--248:16. ACM, 2015.

\bibitem{mardia_jupp_2000}
K.~V. Mardia and P.~E. Jupp.
\newblock {\em Directional statistics}.
\newblock Wiley, 2000.

\bibitem{mohlin2020matrixfisher}
David Mohlin, Josephine Sullivan, and G\'{e}rald Bianchi.
\newblock Probabilistic orientation estimation with matrix fisher
  distributions.
\newblock In {\em Advances in Neural Information Processing Systems},
  volume~33, 2020.

\bibitem{Moon_2020_ECCV_I2L-MeshNet}
Gyeongsik Moon and Kyoung~Mu Lee.
\newblock {I2L-MeshNet}: Image-to-lixel prediction network for accurate {3D}
  human pose and mesh estimation from a single rgb image.
\newblock In {\em Proceedings of the European Conference on Computer Vision
  (ECCV)}, 2020.

\bibitem{Oikarinen2020graphmdn}
Tuomas~P. Oikarinen, Daniel~C. Hannah, and Sohrob Kazerounian.
\newblock {GraphMDN}: Leveraging graph structure and deep learning to solve
  inverse problems.
\newblock {\em CoRR}, abs/2010.13668, 2020.

\bibitem{omran2018nbf}
Mohamed Omran, Christoph Lassner, Gerard Pons-Moll, Peter~V. Gehler, and Bernt
  Schiele.
\newblock Neural body fitting: Unifying deep learning and model-based human
  pose and shape estimation.
\newblock In {\em Proceedings of the International Conference on 3D Vision
  (3DV)}, 2018.

\bibitem{SMPL-X:2019}
Georgios Pavlakos, Vasileios Choutas, Nima Ghorbani, Timo Bolkart, Ahmed A.~A.
  Osman, Dimitrios Tzionas, and Michael~J. Black.
\newblock Expressive body capture: {3D} hands, face, and body from a single
  image.
\newblock In {\em Proceedings of the IEEE Conference on Computer Vision and
  Pattern Recognition (CVPR)}, 2019.

\bibitem{pavlakos2019texturepose}
Georgios Pavlakos, Nikos Kolotouros, and Kostas Daniilidis.
\newblock Texturepose: Supervising human mesh estimation with texture
  consistency.
\newblock In {\em Proceedings of the IEEE International Conference on Computer
  Vision (ICCV)}, 2019.

\bibitem{pavlakos2018humanshape}
Georgios Pavlakos, Luyang Zhu, Xiaowei Zhou, and Kostas Daniilidis.
\newblock Learning to estimate 3{D} human pose and shape from a single color
  image.
\newblock In {\em Proceedings of the IEEE Conference on Computer Vision and
  Pattern Recognition (CVPR)}, 2018.

\bibitem{deepdirectstat2018}
Sergey Prokudin, Peter Gehler, and Sebastian Nowozin.
\newblock Deep directional statistics: Pose estimation with uncertainty
  quantification.
\newblock In {\em Proceedings of the European Conference on Computer Vision
  (ECCV)}, Sept. 2018.

\bibitem{ravi2020pytorch3d}
Nikhila Ravi, Jeremy Reizenstein, David Novotny, Taylor Gordon, Wan-Yen Lo,
  Justin Johnson, and Georgia Gkioxari.
\newblock Accelerating {3D} deep learning with {PyTorch3D}.
\newblock {\em arXiv:2007.08501}, 2020.

\bibitem{Rezende2015normflows}
Danilo Rezende and Shakir Mohamed.
\newblock Variational inference with normalizing flows.
\newblock In Francis Bach and David Blei, editors, {\em Proceedings of the
  International Conference on Machine Learning}, volume~37 of {\em Proceedings
  of Machine Learning Research}, pages 1530--1538, Lille, France, 07--09 Jul
  2015. PMLR.

\bibitem{saito2019pifu}
Shunsuke Saito, Zeng Huang, Ryota Natsume, Shigeo Morishima, Angjoo Kanazawa,
  and Hao Li.
\newblock Pifu: Pixel-aligned implicit function for high-resolution clothed
  human digitization.
\newblock In {\em Proceedings of the IEEE International Conference on Computer
  Vision (ICCV)}, October 2019.

\bibitem{saito2020pifuhd}
Shunsuke Saito, Tomas Simon, Jason Saragih, and Hanbyul Joo.
\newblock Pifuhd: Multi-level pixel-aligned implicit function for
  high-resolution {3D} human digitization.
\newblock In {\em Proceedings of the IEEE Conference on Computer Vision and
  Pattern Recognition (CVPR)}, June 2020.

\bibitem{STRAPS2020BMVC}
Akash Sengupta, Ignas Budvytis, and Roberto Cipolla.
\newblock Synthetic training for accurate {3D} human pose and shape estimation
  in the wild.
\newblock In {\em Proceedings of the British Machine Vision Conference (BMVC)},
  September 2020.

\bibitem{sengupta2021probabilisticposeshape}
Akash Sengupta, Ignas Budvytis, and Roberto Cipolla.
\newblock Probabilistic {3D} human shape and pose estimation from multiple
  unconstrained images in the wild.
\newblock In {\em Proceedings of the IEEE Conference on Computer Vision and
  Pattern Recognition (CVPR)}, 2021.

\bibitem{Sminchisescu2005bm3e}
C. Sminchisescu, A. Kanaujia, Zhiguo Li, and D. Metaxas.
\newblock Discriminative density propagation for {3D} human motion estimation.
\newblock In {\em Proceedings of the IEEE Conference on Computer Vision and
  Pattern Recognition (CVPR)}, volume~1, pages 390--397 vol. 1, 2005.

\bibitem{Sminchisescu2001covsampling}
Cristian Sminchisescu and Bill Trigg.
\newblock Covariance scaled sampling for monocular {3D} body tracking.
\newblock In {\em Proceedings of the IEEE Conference on Computer Vision and
  Pattern Recognition (CVPR)}, 2001.

\bibitem{Sminchisescu2002hyper}
Cristian Sminchisescu and Bill Trigg.
\newblock Hyperdynamics importance sampling.
\newblock In {\em Proceedings of the European Conference on Computer Vision
  (ECCV)}, 2002.

\bibitem{Sminchisescu2003kinematicjump}
Cristian Sminchisescu and Bill Trigg.
\newblock Kinematic jump processes for monocular {3D} human tracking.
\newblock In {\em Proceedings of the IEEE Conference on Computer Vision and
  Pattern Recognition (CVPR)}, 2003.

\bibitem{smith20193dfromsilhouettes}
B.~M. {Smith}, V. {Chari}, A. {Agrawal}, J.~M. {Rehg}, and R. {Sever}.
\newblock Towards accurate {3D} human body reconstruction from silhouettes.
\newblock In {\em Proceedings of the International Conference on 3D Vision
  (3DV)}, 2019.

\bibitem{sun2019hrnet}
Ke Sun, Bin Xiao, Dong Liu, and Jingdong Wang.
\newblock Deep high-resolution representation learning for human pose
  estimation.
\newblock In {\em CVPR}, 2019.

\bibitem{tan2017}
Vince J.~K. Tan, Ignas Budvytis, and Roberto Cipolla.
\newblock Indirect deep structured learning for {3D} human shape and pose
  prediction.
\newblock In {\em Proceedings of the British Machine Vision Conference (BMVC)},
  2017.

\bibitem{varol18_bodynet}
G{\"u}l Varol, Duygu Ceylan, Bryan Russell, Jimei Yang, Ersin Yumer, Ivan
  Laptev, and Cordelia Schmid.
\newblock {BodyNet}: Volumetric inference of {3D} human body shapes.
\newblock In {\em Proceedings of the European Conference on Computer Vision
  (ECCV)}, 2018.

\bibitem{varol17_surreal}
G{\"u}l Varol, Javier Romero, Xavier Martin, Naureen Mahmood, Michael~J. Black,
  Ivan Laptev, and Cordelia Schmid.
\newblock Learning from synthetic humans.
\newblock In {\em Proceedings of the IEEE Conference on Computer Vision and
  Pattern Recognition (CVPR)}, 2017.

\bibitem{vonMarcard2018}
Timo von Marcard, Roberto Henschel, Michael Black, Bodo Rosenhahn, and Gerard
  Pons-Moll.
\newblock Recovering accurate {3D} human pose in the wild using {IMUs} and a
  moving camera.
\newblock In {\em Proceedings of the European Conference on Computer Vision
  (ECCV)}, 2018.

\bibitem{Wehrbein2021posenormflows}
Tom Wehrbein, Marco Rudolph, Bodo Rosenhahn, and Bastian Wandt.
\newblock Probabilistic monocular {3D} human pose estimation with normalizing
  flows.
\newblock In {\em Proceedings of the IEEE International Conference on Computer
  Vision (ICCV)}, 2021.

\bibitem{wu2019detectron2}
Yuxin Wu, Alexander Kirillov, Francisco Massa, Wan-Yen Lo, and Ross Girshick.
\newblock Detectron2.
\newblock \url{https://github.com/facebookresearch/detectron2}, 2019.

\bibitem{Xu_2019_ICCV}
Yuanlu Xu, Song-Chun Zhu, and Tony Tung.
\newblock {DenseRaC}: Joint {3D} pose and shape estimation by dense
  render-and-compare.
\newblock In {\em Proceedings of the IEEE International Conference on Computer
  Vision (ICCV)}, 2019.

\bibitem{yu15lsun}
Fisher Yu, Yinda Zhang, Shuran Song, Ari Seff, and Jianxiong Xiao.
\newblock {LSUN}: Construction of a large-scale image dataset using deep
  learning with humans in the loop.
\newblock {\em arXiv preprint arXiv:1506.03365}, 2015.

\bibitem{Zanfir_2018_CVPR}
Andrei Zanfir, Elisabeta Marinoiu, and Cristian Sminchisescu.
\newblock Monocular {3D} pose and shape estimation of multiple people in
  natural scenes - the importance of multiple scene constraints.
\newblock In {\em Proceedings of the IEEE Conference on Computer Vision and
  Pattern Recognition (CVPR)}, 2018.

\bibitem{Zeng_2020_CVPR_mesh_dense}
Wang Zeng, Wanli Ouyang, Ping Luo, Wentao Liu, and Xiaogang Wang.
\newblock 3d human mesh regression with dense correspondence.
\newblock In {\em Proceedings of the IEEE Conference on Computer Vision and
  Pattern Recognition (CVPR)}, June 2020.

\bibitem{zhang2019danet}
Hongwen Zhang, Jie Cao, Guo Lu, Wanli Ouyang, and Zhenan Sun.
\newblock Danet: Decompose-and-aggregate network for {3D} human shape and pose
  estimation.
\newblock In {\em Proceedings of the 27th ACM International Conference on
  Multimedia}, pages 935--944, 2019.

\end{thebibliography}
}

\end{document}